%% file: main.tex
\documentclass[10pt,twocolumn,letterpaper]{article}

\usepackage{cvpr}              

\input{preamble}

\definecolor{cvprblue}{rgb}{0.21,0.49,0.74}
\usepackage[pagebackref,breaklinks,colorlinks,allcolors=cvprblue]{hyperref}

\def\confName{CVPR}
\def\confYear{2025}

\title{Expanding-and-Shrinking Binary Neural Networks}

\author{
Xulong Shi${^{1}}$ \quad Caiyi Sun${^{2}}$ \quad Zhi Qi${^2}$ \quad Liu Hao${^2}$ \quad Xiaodong Yang${^{1}}$ \\
$^1$QCraft \quad $^2$Southeast University \\
}

\begin{document}
\maketitle
\input{sec/0_abstract}    
\input{sec/1_intro}

\input{sec/2_related}

\input{sec/3_method}
\input{sec/4_experiment}

\input{sec/5_ablation}

\input{sec/6_complexity}

\input{sec/7_conclusion}

{
    \small
    \bibliographystyle{ieeenat_fullname}
    \bibliography{main}
}

\appendix
\input{sec/X_suppl}

\end{document}

%% file: preamble.tex
\usepackage{stackengine} 
\newcommand\oast{\stackMath\mathbin{\stackinset{c}{0ex}{c}{0ex}{\ast}{\bigcirc}}}

\usepackage{multirow}

\usepackage{makecell}

\usepackage{graphicx}
\usepackage{amsmath}
\usepackage{amssymb}
\usepackage{booktabs}
\usepackage{wrapfig}
\usepackage{soul}
\usepackage{comment}

\usepackage{algorithm}
\usepackage{algorithmicx}
\usepackage{algpseudocode}

\newcommand{\fc}[2]{\multirow{1}{*}{\(\begin{array}{c}(\text{#1, #2})\end{array}\)}
}
\newlength\savewidth\newcommand\shline{\noalign{\global\savewidth\arrayrulewidth\global\arrayrulewidth 1pt}\hline\noalign{\global\arrayrulewidth\savewidth}}

\newcommand{\bresblocka}[3]{\multirow{3}{*}{\(\begin{array}{c}\left[\text{3$\times$3, #1, #2}\right]\\[-.1em] \left[\text{3$\times$3, #1,1}\right] $\times$#3 \end{array}\)}
}
\newcommand{\avgpoola}[1]{\multirow{1}{*}{\(\begin{array}{c}\text{#1$\times$#1}\end{array}\)}
}
\newcommand{\convblocka}[2]{\multirow{1}{*}{\(\begin{array}{c}(\text{3$\times$3, #1, #2})\end{array}\)}
}
\usepackage{listings}

\usepackage{xcolor}

%% file: sec/0_abstract.tex
\begin{abstract}

While binary neural networks (BNNs) offer significant benefits in terms of speed, memory and energy, they encounter substantial accuracy degradation in challenging tasks compared to their real-valued counterparts. Due to the binarization of weights and activations, the possible values of each entry in the feature maps generated by BNNs are strongly constrained. To tackle this limitation, we propose the expanding-and-shrinking operation, which enhances binary feature maps with negligible increase of computation complexity, thereby strengthening the representation capacity. Extensive experiments conducted on multiple benchmarks reveal that our approach generalizes well across diverse applications ranging from image classification, object detection to generative diffusion model, while also achieving remarkable improvement over various leading binarization algorithms based on different architectures including both CNNs and Transformers.     

\end{abstract}

%% file: sec/1_intro.tex
\section{Introduction}
\label{sec:intro}

Deep neural networks (DNNs) have achieved tremendous impact in a broad range of domains~\cite{video,qn-mixer,recon,pillarnext,prernn,gedepth}. However, the continuously increasing demand of computation and memory poses significant challenges for deployment on mobile or embedded platforms that support various real-time applications, e.g., robots, augmented reality, and autonomous driving~\cite{ar,x-embodiment,tip}.    

In order to address this issue, numerous methods have been proposed, including quantization~\cite{zhou2016dorefa,gong2019dsq,esser2019lsq}, pruning~\cite{liu2017slimming,liu2019metapruning,li2022randompruning}, knowledge distillation~\cite{hinton2015distilling,yang2022focaldistill,wang2023distillbev}, and compact network design~\cite{howard2017mobilenets,zhang2018shufflenet,liu2018darts}. 
At the extreme end of quantization, binary neural networks (BNNs) emerge as one of the most promising techniques to fulfill the onboard deployment with constrained computation and memory resources. As demonstrated in~\cite{rastegari2016xnor}, BNNs can yield $32\times$ memory compression and up to $58\times$ computation reduction on CPU, and they can be accelerated further on FPGA. In addition, BNNs perform only bit-wise operations using the arithmetic-logic unit, which is also more energy-efficient than the floating-point unit. 

\begin{figure}[t]
    \centering
    \includegraphics[width=\linewidth]{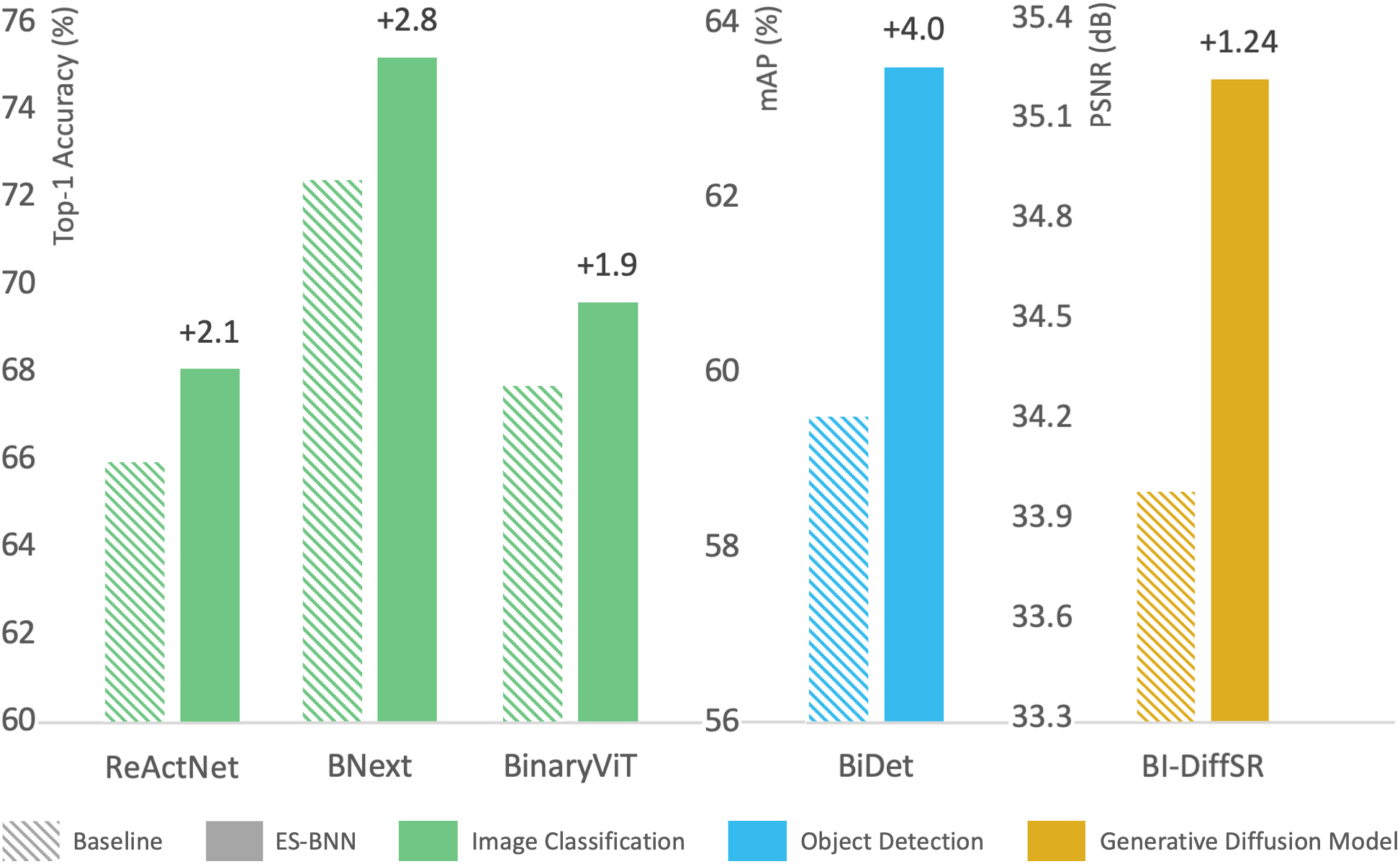}
    \vspace{-2mm}
    \caption{Overview of performance improvement for image classification (top-1 accuracy) on ImageNet, object detection (mAP) on PASCAL VOC, and generative diffusion model based image super-resolution (PSNR) on Manga. Enabled by the proposed ES-BNN, various binary neural networks based on both CNNs and Transformers obtain consistent and significant performance boost.}
    \label{fig:model_improvement}
\end{figure}

In spite of the aforementioned merits in memory, speed and energy, BNNs suffer from large accuracy degradation in challenging tasks. 
To mitigate the gap between BNNs and their real-valued counterparts, prior research has made substantial efforts in optimizing the forward inference and backward propagation algorithms. However, due to the binarization nature of weights and activations (i.e., only two possible values: $-1$ and $1$), each entry of output feature maps produced in BNNs is strongly restricted in a limited number of possible values, and the number is determined by 
the specific architectural configuration of each layer. We define this number as the \textbf{representation capacity} of the corresponding layer, and show that such limited representation capacity is a fundamental issue leading to the accuracy degradation. Regardless of how binarization algorithms are improved, they are ultimately confined by the limited representation capacity.

In light of the above observations, we propose a simple yet effective expanding-and-shrinking operation, which expands input channels (i.e., output channels of a preceding layer) and shrinks convolution kernels. This operation leads to an enhanced representation capacity of output features in each individual layer, while maintaining its original computation complexity and parameter number. Additionally, we combine the expanding-and-shrinking operation with binary group convolution, and their integration results in a notably advantageous synergy: the former inherently enables cross-group information flow, while the latter offers a more favorable balance between expanding and shrinking. We refer to a binary neural network enhanced by the proposed expanding-and-shrinking operation as \textbf{ES-BNN}. Thanks to the proposed generalizable operation, our approach is flexible to fit in diverse binarization algorithms, network architectures, as well as downstream tasks. 
As demonstrated in Figure~\ref{fig:model_improvement}, ES-BNN consistently and remarkably improves multiple representative BNNs. 

Our main contributions are summarized as follows. First, we propose the expanding-and-shrinking operation, which boosts the representation capacity of each binary layer and improves the overall accuracy, at the cost of negligible increase in the computation complexity of an original BNN. 
Second, our approach is applicable to various leading binarization algorithms as well as network architectures from CNNs to Transformers. Third, our approach generalizes in diverse downstream applications ranging from image classification, object detection to generative diffusion model, while rendering superior results on multiple benchmarks. Our code and models are released at \url{https://github.com/imfinethanks/ESBNN}.

%% file: sec/2_related.tex
\section{Related Work}
\label{sec:related}

Since the introduction of binary neural networks, a number of research works are dedicated to minimize the information loss incurred by the binarization of weights and activations. In this field, remarkable progress has been made in optimizing the training and binarization algorithms. 

Binarization of a neural network originates from the pioneering works in~\cite{ebp,courbariaux2016bnns}, which establish the end-to-end trainable framework of binary weights and activations through expectation backpropagation and straight-through estimator, respectively. A series of following methods are proposed to improve the training algorithms. IR-Net~\cite{qin2020irnet} introduces the error decay estimator to mitigate information loss of gradients by gradually approximating the sign function in backward propagation, which guarantees adequate updates at the beginning and accurate gradients at the end of training. RBNN~\cite{lin2020rbnn} develops a training-aware approximation of the sign function to enable gradient propagation. ReCU~\cite{xu2021recu} exploits the rectified clamp unit to revive the dead weights that are barely updated during training. AdamBNN~\cite{liu2021adambnn} analyzes the impact of weight decay and Adam for training, based on which a simple training scheme is derived. BNext~\cite{guo2022bnext} proposes
a diversified consecutive knowledge-distillation technique to alleviate the counter-intuitive overfitting problem.

In parallel, a large family of the binarization research attempts to compensate for the information loss in binarized features. XNOR-Net~\cite{rastegari2016xnor} first uses two types of real-valued scaling factors for both weights and activations in binarization so as to minimize the quantization error. Bi-Real~\cite{liu2018birealnet} adds real-valued shortcuts to propagate real-valued features in order to complement the binarized main branch. Real-to-Bin~\cite{martinez2020real-to-binary} reduces the output discrepancy between the binary and the corresponding real-valued convolution based on the proposed real-to-binary attention matching and data-driven channel re-scaling. Group-Net~\cite{zhuang2019Group-Net} and BENN~\cite{zhu2019benn} both leverage the ensemble methods to combine multiple binary models to obtain performance gains. ReActNet~\cite{liu2020reactnet} introduces RSign and RPRelu to explicitly shift and reshape activation distribution, and uses a distributional loss to further align features between binary and real-valued networks. AdaBin~\cite{tu2022adabin} adaptively obtains an optimal binary set of weights and activations for each layer instead of a fixed set. INSTA-BNN~\cite{lee2023insta} dynamically controls the quantization threshold in an instance-aware manner based on the representative higher-order statistics to estimate the characteristics of input distribution.

%% file: sec/3_method.tex
\section{Method}
\label{sec:method}

In this section, we start from providing an overview, which includes the fundamental principles of binarization and the underlying motivation of our approach. We then present the proposed expanding-and-shrinking operation as well as the synergistic binary group convolution.

\begin{figure*}[t]
\centering
\includegraphics[width=1.0\textwidth]{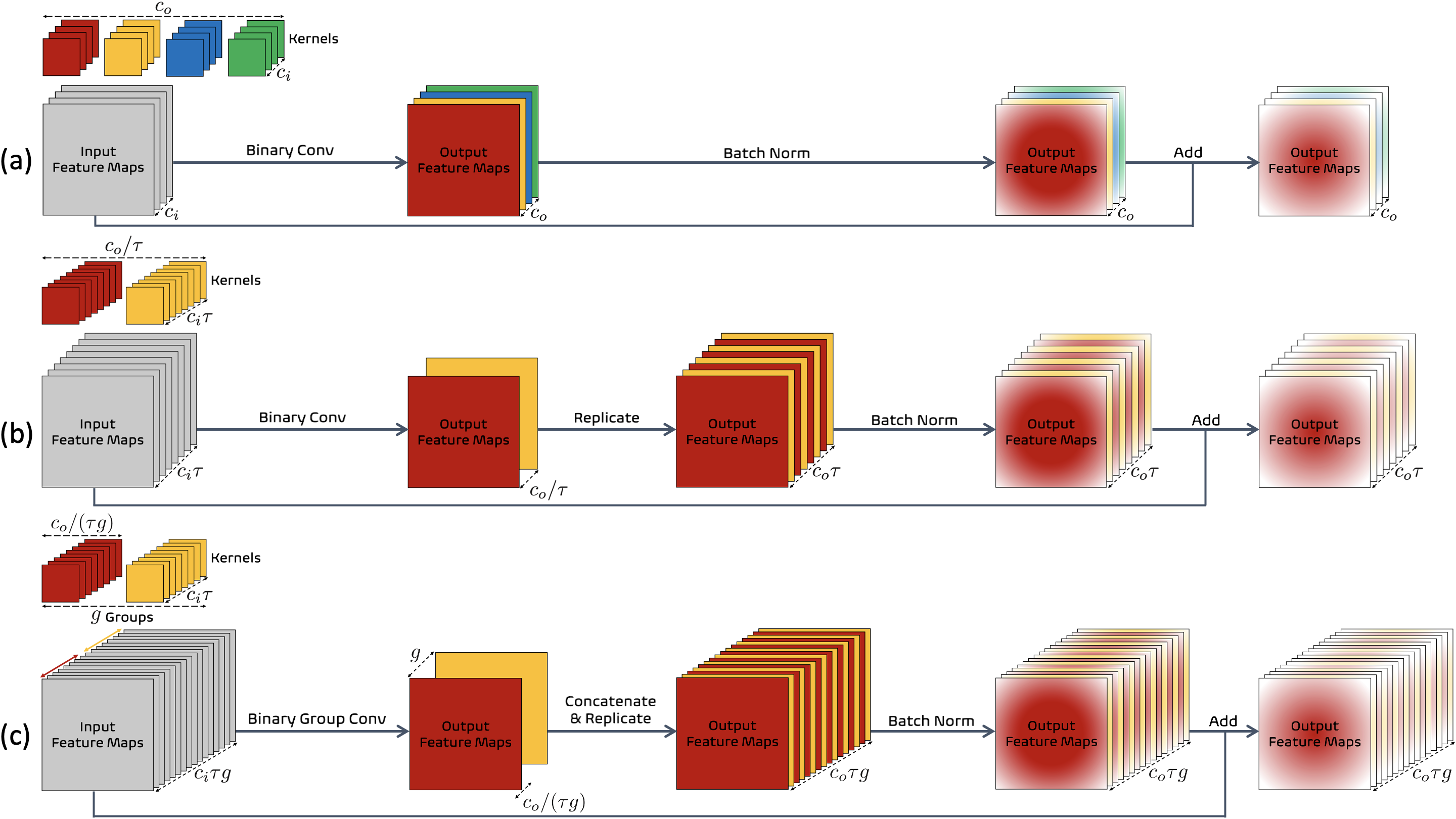}
\caption{A schematic overview of the proposed approach ES-BNN utilized in each individual layer of a binary neural network. (a) shows a standard binary convolution layer. (b) illustrates the expanding-and-shrinking operation applied on top of (a). In (c) the binary group convolution is further integrated based on (b).}
\label{fig:overview}
\end{figure*}

\subsection{Problem Statement}
In the following, we use CNNs to illustrate the formulation of our approach, which is also straightforward to be applied to Transformers.
We denote by $W \in \mathbb{R}^{c_{o} \times c_{i} \times k \times k}$ and $A \in \mathbb{R}^{c_{i} \times w \times h}$ the weights and input features of a convolutional layer, where $c_o$ and $c_i$ represent the number of output and input channels, $w$ and $h$ indicate the width and height of the input features, and $k$ is the spatial dimension of the kernel. In a BNN, both weights and features are binarized, and the convolution is performed as:
\begin{equation}
A * W \approx \mathcal{B}_A(A) \oast \mathcal{B}_W(W), 
\label{eq-1}
\end{equation}
where $\mathcal{B}_A$ and $\mathcal{B}_W$ represent the binarization functions of weights and features, and $\oast$ is the binary convolution that can be implemented by efficient bit-wise operations. While the specific algorithms of forward inference and backward propogation of $\mathcal{B}_A$ and $\mathcal{B}_W$ can be different in various methods, $\mathcal{B}_A(A)$ and $\mathcal{B}_W(W)$ have only two possible values (i.e., $-1$ and $+1$) represented by a single bit.

Thus, unlike the real-valued convolution, the output of binary convolution is constrained in a limited set of values, and the entry at $[i, x, y]$ of an output feature map is: 
\begin{equation*}
\resizebox{\linewidth}{!}{
$\sum\limits_j\sum\limits_{dx}\sum\limits_{dy}\mathcal{B}_{A}(A[j,x+dx,y+dy])\mathcal{B}_{W}(W[i,j,x+dx,y+dy]),$
}
\end{equation*}
where $j = 0, ..., c_i - 1$, and $dx, dy = -(k - 1)/2, ..., (k - 1)/2$. Therefore, the value of each entry of the output feature map is in the range of $[-c_i k^2$, $c_i k^2]$ with an interval of 2 (i.e., there are $c_i k^2 + 1$ possible values in total). We define this total number of possible values as the \textbf{representation capacity} of the corresponding layer in a BNN. 

As aforementioned, research works in this field mostly focus on improving the forward inference and backward propogation algorithms to be better tailored for binarization. 
However, as long as $A$ and $W$ in Equation~\ref{eq-1} undergo binarization, their output feature maps are restricted by the representation capacity, which strongly delimits the overall accuracy of a BNN.   

In this paper, we seek to answer the research question: \textit{How can we push the boundary of representation capacity, and meanwhile, maintain the original computation complexity of a BNN?}

Directly increasing $c_i$ or $k$ indeed escalates the representation capacity, but incurs dramatic growth in the computation complexity as well. 
In order to overcome this dilemma, we propose the expanding-and-shrinking operation, which strengthens the representation capacity, and consequently, improves the final accuracy of various BNNs, with negligible extra computation overhead.

\subsection{Expanding and Shrinking}
\label{sec:expanding-and-shrinking}

\begin{figure*}[t]
\centering
\includegraphics[width=\linewidth]{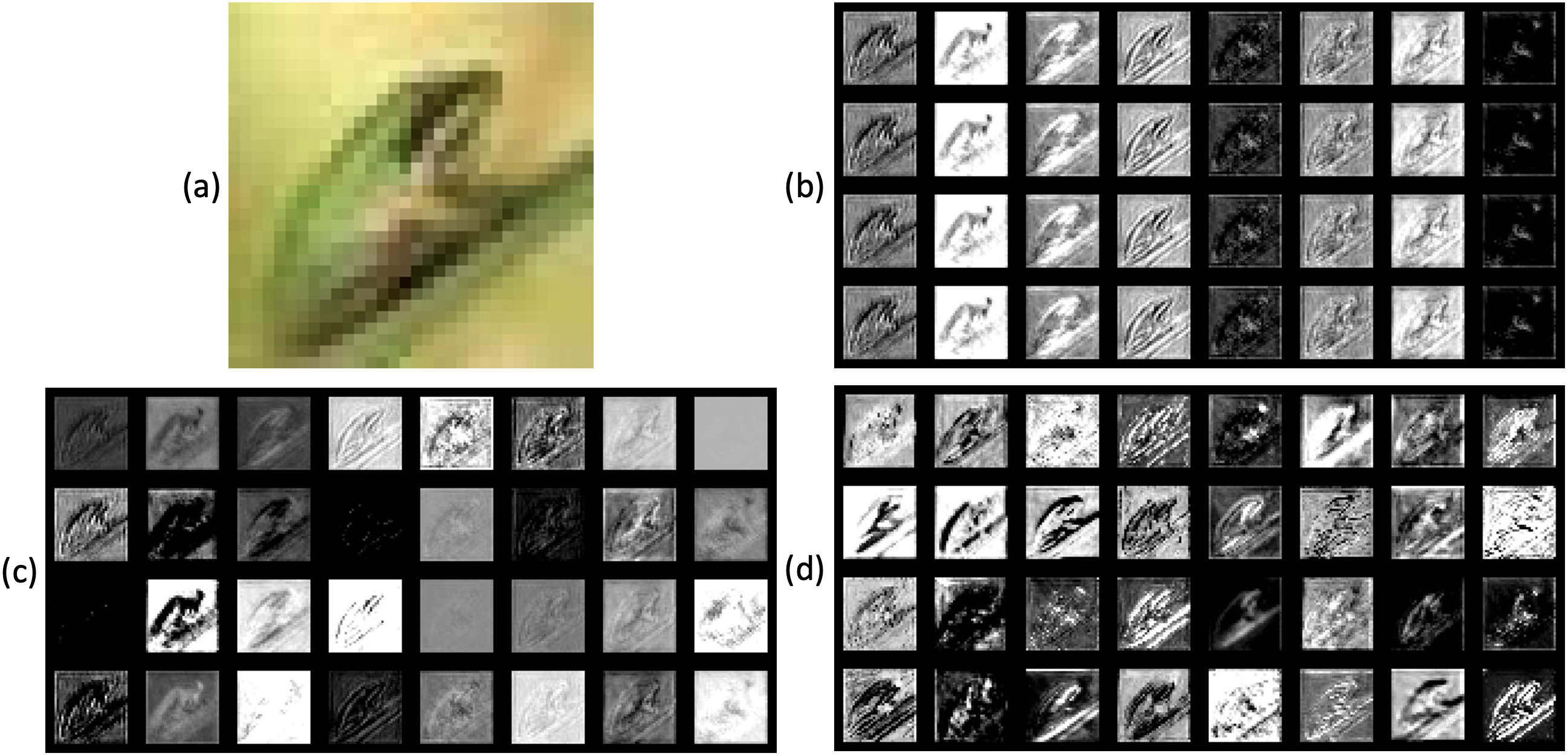}
\vspace{-2mm}
\caption{Illustration of the diversification process of the replicated feature maps in the proposed approach. (a) shows the input image. (b) presents the replication of 8 feature maps (in one row) from the 7th layer in ResNet-20. (c) demonstrates the feature maps in (b) after the batch normalization. (d) corresponds to the feature maps in (c) after the residual connection.}
\label{fig:diversify}  
\end{figure*}

Figure~\ref{fig:overview}(a-b) illustrates the comparison between a standard binary convolution and our proposed approach. We first expand the input channels from $c_i$ to $c_i \tau$, where $\tau \in \mathbb{Z}^+$ is a scaling factor to control the increase of representation capacity. We then shrink the convolution kernels from $c_o$ to $c_o / \tau$ with the purpose of maintaining a constant computation complexity. This is equivalent to reshape $A$ to $\widetilde{A} \in \mathbb{R}^{c_{i} \tau \times w \times h}$ and $W$ to $\widetilde{W} \in \mathbb{R}^{c_{o} / \tau \times c_{i} \tau \times k \times k}$. However, such shrinked output feature maps undermine the representation capacity of next layer (i.e., the number of input channels for next layer is also reduced), and can be incompatible with the architectural unit (e.g., the numbers of feature channels are required to match in a residual connection). To compensate for the shrinking, we expand the output channels by making $\tau^2$ copies of $\mathcal{B}_A(\widetilde{A}) \oast \mathcal{B}_W(\widetilde{W})$ and concatenating them along the channel dimension, such that the final number of output channels of current layer become $c_o \tau$, which essentially makes the input channels of next layer expanded. In this manner, we can continuously enhance the representation capacity across all layers through the expanding-and-shrinking operation, at the same time, preserve the original computation complexity.

Next we discuss how the replicated feature maps contribute to the final feature representation of each layer in the proposed approach. In a typical BNN with batch normalization and residual connection, the computation process of a single layer in our ES-BNN can be expressed as:    
\begin{align}
\begin{split}
X &= \texttt{Rep}\left(\tau^2, \mathcal{B}_A(\widetilde{A}) \oast \mathcal{B}_W(\widetilde{W})\right), \\
Y &= \gamma \odot \frac{X - \mu_X}{\sigma_X} + \beta + \widetilde{Y},
\end{split}
\label{eq:bn}
\end{align}
where $X$ is the expanded feature maps through replication, $\mu_X$, $\sigma_X$, $\gamma$ and $\beta$ are respectively the mean, variance, scale and shift used to perform batch normalization, and $\widetilde{Y}$ denotes a residual connection. 
As we can see, the channel-wise parametric $\gamma$ and $\beta$ first make the replicated feature maps different. In addition, $\widetilde{Y}$ that accumulates the differences from previous layers diversifies these feature maps further. 
Therefore, a standard binary convolutional layer modified by our expanding-and-shrinking operation is still able to produce diverse output feature maps.   
In Figure~\ref{fig:diversify}, we illustrate this diversification process. It is apparent to observe that the replicated feature maps based on the expanding-and-shrinking operation turn into heterogeneous after batch normalization and residual connection. 

\subsection{Binary Group Convolution}
\label{sec:group_conv}
Increasing $\tau$ expands the representation capacity, but it also shrinks the convolution kernels, although the overall numbers of learnable parameters in $W$ and $\widetilde{W}$ remain the same. In the extreme case (i.e., $\tau = c_o$), there would be only one output channel generated directly by the binary convolution, 
undermining the diversity of output features. 
This trade-off between expanding and shrinking is indeed observed in our experiments (see Section~\ref{sec:ablation} and Table~\ref{t_g_effect}). 

In order to better balance expanding and shrinking, we introduce binary group convolution, which provides different input features and distinct batch normalization in each group, thus infusing greater diversity into the output features. As illustrated in Figure~\ref{fig:overview}(c), given $g$ groups, we start from expanding the input channels to $c_i \tau g$, and distribute them to the $g$-group binary convolutions, each of which takes the expanding-and-shrinking operation as described above. In a single group, to maintain computation complexity, $c_o / (\tau g)$ output channels are generated. After concatenating the output features from $g$ groups, the $c_o / \tau$ output channels are in the end expanded to $c_o \tau g$, which readies the input channels for the next binary group convolution layer. 

Group convolution is traditionally used together with the point-wise convolution~\cite{howard2017mobilenets} or the channel shuffle operation~\cite{zhang2018shufflenet} due to the lack of interaction between channels in different groups. However, our binary group convolution is exempted from such a complication, and enables cross-group information flow between multiple groups by nature. This is realized by the shrinking operation that condenses all channel information into $c_o / \tau$ output channels collected from all groups. And each group in next layer receives $\tau^2$ copies of $c_o / \tau$ channels (as detailed in Equation~\ref{eq:bn}), facilitating the exchange of information across groups.

\subsection{Implementation}

In practice, the scaling factor $\tau$ in expanding and shrinking, as well as the group number $g$ in binary group convolution, are not necessarily the same across different layers. Following the general principle in network architecture design (i.e., deeper layers with increased channels for feature abstraction), we adopt larger $\tau g$ in deeper layers to increase the representation capacity more. See more details in the appendix.  

Apart from CNNs, our approach is also applicable to Transformers, where the feed-forward network (FFN) can be viewed as a point-wise convolution. This allows for the straightforward application of the proposed operation as described above. However, this results in substantially higher computation complexity in the self- or cross-attention layer due to the increased channels. We thus average the features of increased channels to restore to the original channel number before forwarding to an attention layer so as to preserve the initial complexity of attention. More details can be found in the appendix.  

%% file: sec/4_experiment.tex
\section{Experiments}

In this section, we first describe the experimental setup, and then report extensive comparisons with the state-of-the-art methods on the popular benchmarks. A wide range of ablation study and related analysis are provided for in-depth understanding of the proposed approach.  

\subsection{Experimental Setup}

\noindent\textbf{Datasets.} 
To comprehensively evaluate ES-BNN, we perform extensive experiments on a broad range of applications including image classification, object detection, as well as generative diffusion model based image super-resolution. These tasks are widely evaluated on seven benchmarks including CIFAR-10~\cite{krizhevsky2009cifar10}, ImageNet~\cite{russakovsky2015imagenet}, PASCAL VOC~\cite{mark2010voc}, Set5~\cite{martin2001database}, B100~\cite{huang2015single}, Urban100~\cite{huang2015single} and Manga109~\cite{matsui2017sketch}. We follow the standard experimental protocols for fair comparisons with previous methods. We provide details of the multiple datasets in the appendix.   

\noindent\textbf{Implementation Details.} 
Thanks to the flexibility of ES-BNN, all experiments are conducted by integrating the proposed operation into the official open-source codebase of original methods, with no changes to the original training settings. As for the real-valued parts in a network (e.g., the first convolution layer and the last fully-connected layer), we take different strategies: for the computationally intensive convolution layer, the expanding-and-shrinking operation is applied to maintain its computation complexity; while for the computationally lightweight fully-connected layer, the channels are modified directly with negligible increase in its computation complexity. More implementation details can be found in the appendix.

\begin{table}[t]
  \centering
  \setlength{\tabcolsep}{5pt} 
  \resizebox{\linewidth}{!}{
  \begin{tabular}{l|l|cc}
    \toprule
                 &Method     & Baseline (Top-1)        &  ES-BNN (Top-1) \\
    \midrule \midrule
    \multirow{6}{*}{\makecell{ResNet-18}}     &Bi-Real~\cite{liu2018birealnet}  &89.7              & \textbf{90.6} (\textbf{+0.9})     \\ 
                 &IR-Net~\cite{qin2020irnet}       & 91.5                  & \textbf{92.9} (\textbf{+1.4}) \\
                 &ReCU~\cite{xu2021recu}         & 92.8                & \textbf{93.9} (\textbf{+1.1})     \\
                 &RBNN~\cite{lin2020rbnn}         & 92.2                  & \textbf{93.4} (\textbf{+1.2})  \\
                 &SiMaN~\cite{lin2022siman}        & 92.5                 & \textbf{92.8} (\textbf{+0.3})        \\   
                 &AdaBin~\cite{tu2022adabin}        & 93.1                & \textbf{93.5} (\textbf{+0.4})  \\   

    \midrule
    \multirow{6}{*}{\makecell{ResNet-20}}     &Bi-Real~\cite{liu2018birealnet}  & 81.4          & \textbf{85.9} (\textbf{+4.5})       \\ 
                 &IR-Net~\cite{qin2020irnet}       & 86.5              & \textbf{89.7} (\textbf{+3.2})\\
                 &ReCU~\cite{xu2021recu}         & 87.4                & \textbf{90.0} (\textbf{+2.6})    \\
                 &RBNN~\cite{lin2020rbnn}         & 87.8                & \textbf{90.4} (\textbf{+2.6})   \\
                 &SiMaN~\cite{lin2022siman}         & 87.4                & \textbf{89.7} (\textbf{+2.3})       \\   
                 &AdaBin~\cite{tu2022adabin}        & 88.2                 & \textbf{89.6} (\textbf{+1.4})       \\   
 
    \bottomrule
  \end{tabular}
  }
  \caption{Comparison of the top-1 accuracy (\%) of baseline methods and our approach using the backbones of ResNet-18 and ResNet-20 on CIFAR-10.}
  \label{t_cifar10}
  \vspace{-1mm}
\end{table}

\subsection{Resutls on Image Classification}

\noindent\textbf{CIFAR-10.} 
To verify the generalizability of the proposed ES-BNN to different BNNs, we first experiment on CIFAR-10 and broadly evaluate with a series of methods, including Bi-Real~\cite{liu2018birealnet}, IR-Net~\cite{qin2020irnet}, RBNN~\cite{lin2020rbnn}, ReCU~\cite{xu2021recu}, SiMaN~\cite{lin2022siman}, and AdaBin\cite{tu2022adabin}. These methods are used as the baselines, and the proposed approach is universally applied on them. As shown in Table~\ref{t_cifar10}, ES-BNN consistently and considerably boosts the top-1 accuracy over all baselines. In particular, the most significant gains are obtained with 1.4\% and 4.5\% based on the backbones of ResNet-18 and ResNet-20, respectively.

\begin{table*}[t]
  \small 
  \centering
  \begin{tabular}{l|l|ccc|c}
    \toprule
                 &Method  &BOPs ($10^9$)    &FLOPs ($10^8$) &OPs ($10^8$)  & Top-1 (\%)          \\
    \midrule \midrule
    \multirow{10}{*}{\makecell{ResNet-18}}    &BNNs~\cite{courbariaux2016bnns} &1.70 &1.33 &1.60   &42.2              \\  
                &XNOR-Net~\cite{rastegari2016xnor} &1.70 &1.33 &1.60   &51.2               \\  
                &Bi-Real~\cite{liu2018birealnet}  &1.68 &1.39 &1.65  &56.4               \\  
                &XNOR-Net++~\cite{bulat2019xnor++} &1.70 &1.33 &1.60   &57.1               \\  
                &IR-Net~\cite{qin2020irnet}  &1.68 &1.37 &1.64  &58.1               \\  
                &RBNN~\cite{lin2020rbnn}  &1.68 &1.37 &1.64  &59.9               \\  
                &ReSTE~\cite{wu2023reste}  &1.68 &1.37 &1.64  &60.9              \\  
                &ReCU~\cite{xu2021recu}  &1.68 &1.37 &1.64  &61.0               \\ 
                 &\textbf{ReCU (ES-BNN)}  &1.68 &1.41 &1.68  &\textbf{64.4} (\textbf{+3.4})              \\ 
                 &ReActNet~\cite{liu2020reactnet}     &1.68 &1.37 &1.64      &65.9          \\
                 &\textbf{ReActNet (ES-BNN)}       &1.68 &1.41  &1.68    &\textbf{68.0} (\textbf{+2.1})      \\ 
    \midrule
    \multirow{7}{*}{\makecell{Customized CNNs}}            &MeliusNet~\cite{bethge2020meliusnet}  &4.62 &1.35 &2.08  &63.6         \\ 
                &Real2Binary~\cite{martinez2020real-to-binary}  &1.68 &1.56 &1.83  &65.4         \\  
          &ReActNet~\cite{liu2020reactnet} &4.82 &0.12 &0.87  &69.4         \\ 
              &AdamBNN~\cite{liu2021adambnn} &4.82 &0.12 &0.87     &70.5      \\
              &INSTA-BNN~\cite{lee2023insta} &4.82 &0.20 &0.95    &71.7      \\
            &BNext~\cite{guo2022bnext}    &4.82 &0.13 &0.88      &72.4             \\
                 &\textbf{BNext (ES-BNN)}   &4.82 &0.14 &0.89   &\textbf{75.2} (\textbf{+2.8})           \\ 
    \midrule
    \multirow{2}{*}{\makecell{Transformers}}     &BinaryViT~\cite{le2023binaryvit}  &3.83 &0.19 &0.79  &67.7         \\ 
                 &\textbf{BinaryViT (ES-BNN)}   &3.83 &0.20 &0.80   &\textbf{69.6} (\textbf{+1.9})      \\  
    \bottomrule
  \end{tabular}
    \caption{Comparison of the top-1 accuracy and the computation complexity (BOPs, FLOPs and OPs) of the state-of-the-art methods and our approach using ResNet-18, customized CNNs and Transformers on ImageNet.}
    \vspace{-3mm}
  \label{t_imagenet}
\end{table*} 

\begin{table}[t]
    \small 
  \centering
  \setlength{\tabcolsep}{5pt} 
  \begin{tabular}{l|l|c}
    \toprule
                 &Method     &mAP (\%)          \\
    \midrule \midrule
    \multirow{6}{*}{\makecell{Faster R-CNN\\(ResNet-18)}}           &BNNs~\cite{courbariaux2016bnns}  & 35.6               \\
                    &XNOR-Net~\cite{rastegari2016xnor}  & 48.4               \\
                &Bi-Real~\cite{liu2018birealnet}  & 58.2               \\  
         &BiDet~\cite{wang2020bidet}  & 59.5          \\  
         &AutoBiDet~\cite{wang2021autobidet}  & 60.4          \\  
                 &\textbf{BiDet (ES-BNN)}         & \textbf{63.5} (\textbf{+4.0})      \\ 

    \bottomrule
  \end{tabular}
    \caption{Comparison of the object detection results of various methods and our approach based on Faster R-CNN using ResNet-18 on PASCAL VOC.}
  \label{t_voc}
\end{table}

\begin{figure}[t]
  \centering
    \includegraphics[width=\linewidth]{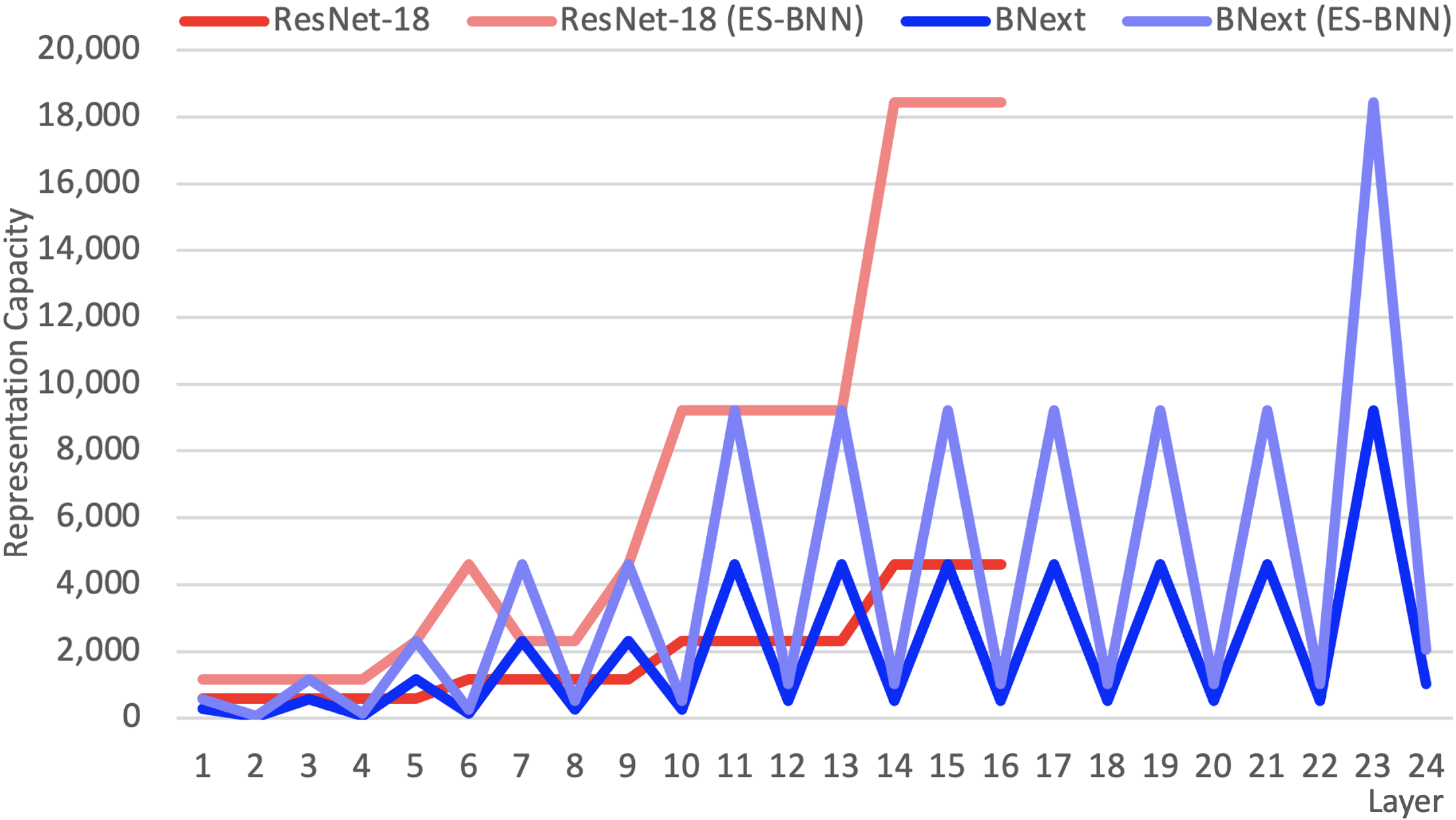}
    \vspace{-5mm}
    \caption{Comparison of the representation capacity of each individual layer before and after applying ES-BNN in a standard architecture ResNet-18 and a customized architecture BNext.}
    \label{fig:rep_cap}
\end{figure}

\noindent\textbf{ImageNet.} 
To validate the proposed approach on the more challenging task, we perform the large-scale image classification on ImageNet. ES-BNN is applied upon a set of representative methods including ReCU~\cite{xu2021recu}, ReActNet~\cite{liu2020reactnet}, BNext~\cite{guo2022bnext}, and BinaryViT~\cite{le2023binaryvit}. As shown in Table~\ref{t_imagenet}, ES-BNN achieves superior results on both ResNet-18 and the customized CNN models. Our approach improves various baseline methods by a clear margin, while maintaining the computation complexity. Taking a closer look into the improvement, ES-BNN delivers a remarkable gain of 2.8\% even on the high-performing method BNext, revealing that our enhancing effect is not diminishing with the strong baseline. 
Furthermore, we evaluate the proposed approach with BinaryViT~\cite{le2023binaryvit}, which is the recent binarization work on Transformers. It is found that ES-BNN is also able to render consistent and significant improvement, suggesting that ES-BNN generalizes for heterogeneous architectures from CNNs to Transformers.  
Figure~\ref{fig:rep_cap} quantitatively exemplifies how the representation capacity of each layer is enhanced by our approach in both standard (ResNet-18) and customized (BNext) network architectures.

\subsection{Results on Object Detection}
\noindent\textbf{PASCAL VOC.}  
Here we assess the generalization of ES-BNN on the task of object detection based on the dataset PASCAL VOC. BiDet~\cite{wang2020bidet} is used as the representative baseline method. As demonstrated in Table~\ref{t_voc}, our approach improves the baseline by 4.0\% in mAP, and largely outperforms other competing algorithms. This experiment further verifies that ES-BNN is capable of adapting to different tasks from classification to detection, showcasing its efficacy in diverse downstream applications.

\begin{table*}[t]
\resizebox{\linewidth}{!}{
\begin{tabular}{l|lll|lll|lll|lll}
\toprule
 & \multicolumn{3}{c|}{Set5}                          & \multicolumn{3}{c|}{B100}                          & \multicolumn{3}{c|}{Urban100}                      & \multicolumn{3}{c}{Manga109}                                                                  \\ 
                   & PSNR                         & SSIM                          & LPIPS                         & PSNR                         & SSIM                          & LPIPS                         & PSNR                         & SSIM                          & LPIPS                         & PSNR                         & SSIM                          & LPIPS                         \\ \midrule \midrule
BNN~\cite{courbariaux2016bnns}                & 13.97          & 0.5210           & 0.4529          & 13.73          & 0.4553          & 0.5784          & 12.75          & 0.4236          & 0.5575          & 9.290           & 0.3035          & 0.7489          \\
DoReFa~\cite{zhou2016dorefa}              & 16.43          & 0.6553          & 0.2662          & 16.11          & 0.5912          & 0.3972          & 15.09          & 0.5495          & 0.4055          & 12.35          & 0.4609          & 0.5047          \\
XNOR~\cite{rastegari2016xnor}                & 32.34          & 0.8661          & 0.0782          & 27.94          & 0.7548          & 0.1665          & 27.47          & 0.8225          & 0.1153          & 31.99          & 0.9428          & 0.0326          \\
IRNet~\cite{qin2020irnet}               & 32.55          & 0.9340           & 0.0446          & 27.76          & 0.8199          & 0.1115          & 26.34          & 0.8452          & 0.0913          & 23.89          & 0.7621          & 0.1820           \\
ReActNet~\cite{liu2020reactnet}            & 34.30           & 0.9271          & 0.0351          & 28.36          & 0.8158          & 0.0943          & 27.43          & 0.8563          & 0.0731          & 32.16          & 0.9441          & 0.0379          \\
BBCU~\cite{xia2022bbcu}                & 34.31          & 0.9281          & 0.0393          & 28.39          & 0.8202          & 0.0905          & 28.05          & 0.8669          & 0.0620           & 32.88          & 0.9508          & 0.0272          \\
BI-DiffSR~\cite{chen2024binarized}           & 35.68          & 0.9414          & \textbf{0.0277} & 29.73          & 0.8478          & \textbf{0.0682} & 28.97          & 0.8815          & 0.0522          & 33.99          & 0.9601          & 0.0172          \\
BI-DiffSR (ES-BNN) & \textbf{36.35} & \textbf{0.9501} & 0.0336          & \textbf{30.17} & \textbf{0.8623} & 0.0776          & \textbf{29.48} & \textbf{0.8934} & \textbf{0.0501} & \textbf{35.23} & \textbf{0.9648} & \textbf{0.0161}  \\  \bottomrule
\end{tabular}}
    \caption{Comparison of generative diffusion model based image super-resolution under three evaluation metrics (PSNR, SSIM and LPIPS) across four benchmark datasets (Set5, B100, Urban100 and Manga109).}
  \label{t_diff}
\end{table*}

\subsection{Results on Generative Diffusion Model}
We adopt image super-resolution (SR) to verify the effectiveness of our approach on generative diffusion model. Due to the high-quality generation performance, the diffusion models~\cite{dm1, dm2} have been widely used in conditional image generation tasks, including SR. However, achieving desirable results with diffusion models requires thousands of iterative steps, resulting in slow inference time. Binarization is therefore well-suited for this application and holds the potential to significantly accelerate diffusion models. We use the most recent method BI-DiffSR~\cite{chen2024binarized} as our baseline. Following the previous works, we employ DIV2K~\cite{timofte2017ntire} and Flickr2K~\cite{bevilacqua2012low} as the training set and evaluate on four benchmark datasets. We report the results of the $\times$2 upscale factor in Table~\ref{t_diff}. As can be seen, our approach delivers superior results under most metrics across the four datasets. This application further demonstrates the versatility of ES-BNN, showing its applicability not only to high-level vision tasks like classification and detection but also to generative low-level vision tasks.

\begin{figure}[t]
    \centering
    \includegraphics[width=\linewidth]{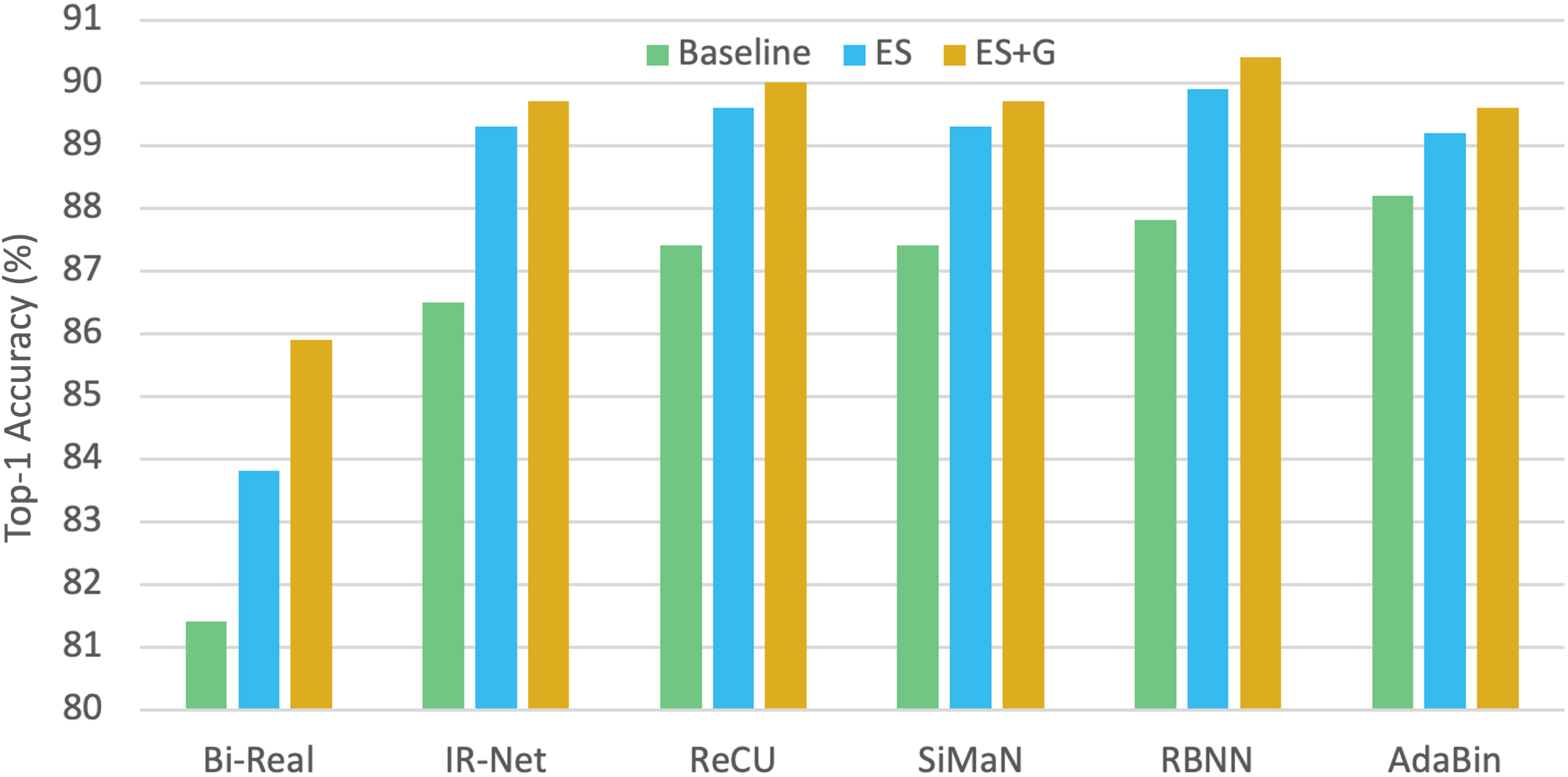}
    \vspace{-3mm}
    \caption{Comparison of our approach with different combinations of the proposed expanding-and-shrinking operation (ES) as well as the binary group convolution (G) using ResNet-20 on CIFAR-10.}
    \label{fig:component_improvement}
\end{figure}

%% file: sec/5_ablation.tex
\subsection{Ablation Study and Analysis}
\label{sec:ablation}

\begin{figure*}[t]

    \centering
    \includegraphics[width=\textwidth]{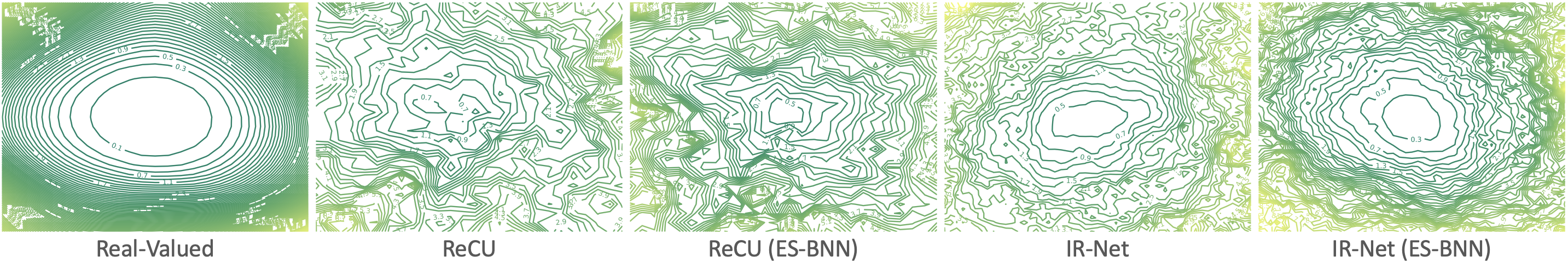}
    \vspace{-4mm}
    \caption{Comparison of the loss landscape (i.e., the contour line view) of baseline methods and our approach using ResNet-20 on CIFAR-10. A real-valued counterpart is demonstrated here for reference.}
    \label{fig:landscape}
\end{figure*}

\noindent\textbf{Contribution of Each Component.} 
Here we investigate the contribution of each individual component in our approach through a set of combinatorial experiments. As shown in Figure~\ref{fig:component_improvement}, starting from the baseline methods, we incrementally integrate with the expanding-and-shrinking operation (i.e., ES) and the binary group convolution (i.e., ES+G). It can be found that the two components consistently make improvement over all baseline methods, which validate the effectiveness of the proposed design.

\begin{table}[t]
\small
  \centering
  \setlength{\tabcolsep}{4pt} 
\begin{tabular}{l|ccc|ccc}
\toprule
\multicolumn{1}{l|}{\multirow{2}{*}{}}&\multicolumn{3}{c|}{$\tau$ $(g=1)$} &\multicolumn{3}{c}{$g$ $(\tau=4)$}   \\ 
\multicolumn{1}{l|}{}  & Bi-Real & IR-Net  & ReCU  & Bi-Real & IR-Net  & ReCU   \\ \midrule\midrule
1 &81.4 &86.5 &87.5 &83.8 &89.4 &89.6 \\
2 &83.4 &88.7 &89.0 &85.2 &89.7 &89.9 \\
4 &83.8 &89.4 &89.6 &85.9 &89.6 &90.0 \\
8 &82.8 &89.3 &89.3 &- &- &- \\
16 &80.2 &88.0 &88.6 &- &- &- \\ \bottomrule
\end{tabular}
\caption{Analysis of the hyper-parameters $\tau$ (scaling factor) and $g$ (group number) in ES-BNN. We report the top-1 accuracy (\%) using ResNet-20 on CIFAR-10.}
\label{t_g_effect}
\end{table}

\noindent\textbf{Hyper-Parameters.} 
We study the two hyper-parameters in our approach: the scaling factor $\tau$ and the group number $g$. 
To first eliminate other impacts, a uniform value of $\tau$ is used throughout the whole network, and the binary group convolution is not utilized. In Table~\ref{t_g_effect}, we observe significant improvement in the accuracy when $\tau$ increases from 1 to 4, showing the effect of expanding representation capacity. However, the accuracy reaches a plateau or drops as $\tau$ increases further, indicating that the effect of shrinking convolution kernels becomes more prominent. Based on the optimal $\tau = 4$, combining binary group convolution can further improve the accuracy. For instance, $\tau = 4$ and $g = 4$ outperforms $\tau = 16$ and $g = 1$, although the overall expanded channels are the same in the two settings. 
Note the maximum value of $g$ is 4, constrained by the limit of output channels in ResNet-20 (i.e., $c_o / (\tau g) \ge 1$).

\begin{figure}
  \centering
    \includegraphics[width=\linewidth]{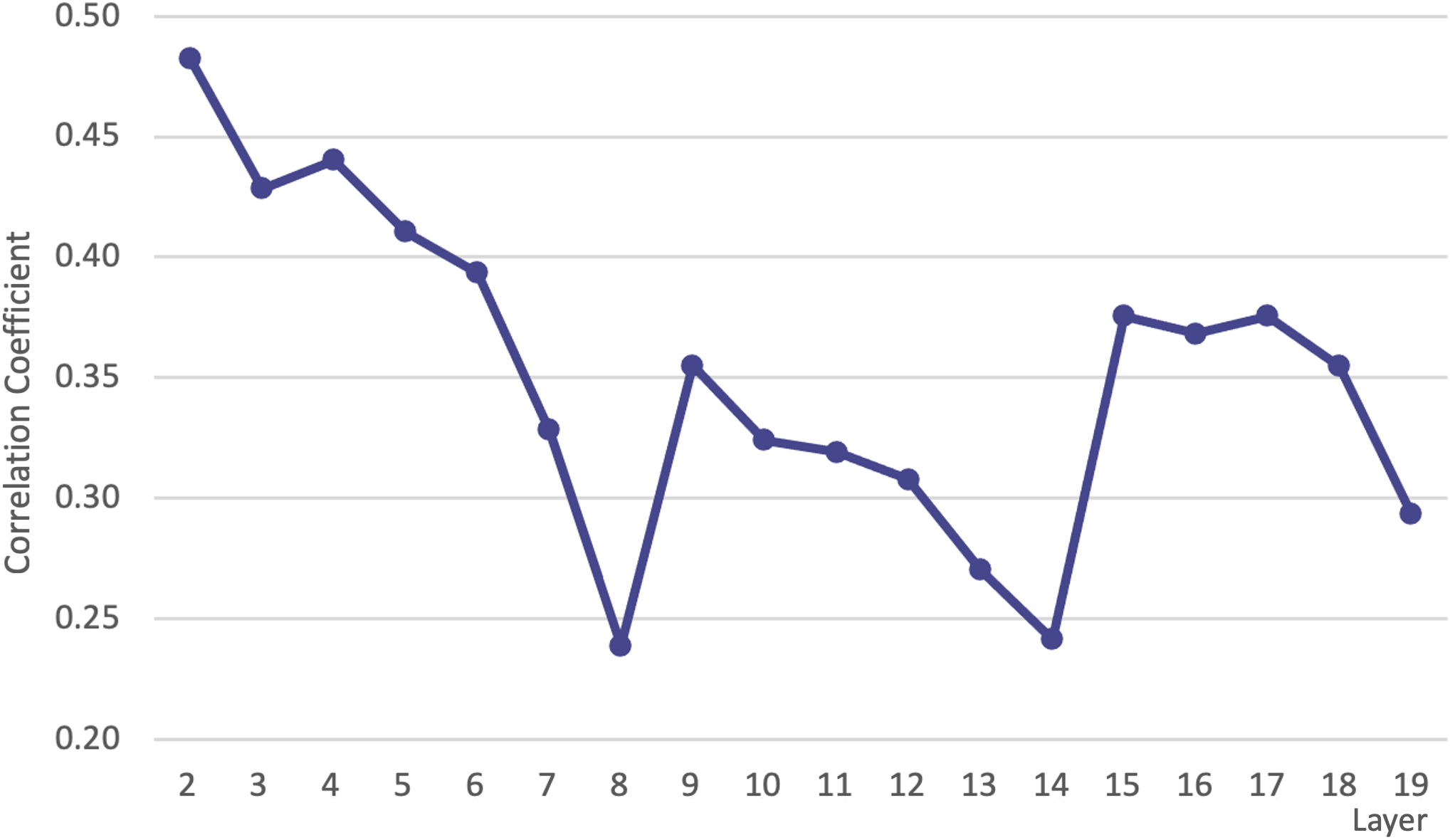}
    \caption{Quantization of the diversification of replicated features by computing the correlation coefficients of binarized layers. We report the result based on ReCU with ResNet-20 on CIFAR-10.} 
    \label{fig:correlation}
\vspace{-5mm}
\end{figure}

\noindent\textbf{Diversification of Replicated Features.} 
In addition to the qualitative visualization of the diversification process in Figure~\ref{fig:diversify}, we provide quantitative evaluation of the diversification of replicated features. We make use of the correlation coefficient to measure the relationship strength between the replicated features. Specifically, we compute the absolute values of correlation coefficients between pairwise feature maps, which include the feature map directly generated by binary convolution and the one replicated from this feature map (see Figure~\ref{fig:overview}(b)). We then take the average of all pairs as the correlation coefficient of the corresponding layer. Figure~\ref{fig:correlation} plots the coefficients of all layers (apart from the first and last layers that are not binarized) in ResNet-20. Initially, the coefficients of all layers are 1. As can be seen from this figure, the coefficients are largely reduced from the initial values, clearly indicating the replicated features are diversified. It can be also observed that the coefficients generally decrease with an increase in layers. Notably, abrupt shifts occur at the 9th and the 15th layers, aligning with the architectural variations of ResNet-20 (i.e., zero-padding in feature channels for residual connections, see more details in the appendix).

\noindent\textbf{Loss Landscape.} 
For the in-depth understanding of the improvement achieved through our approach, we further visualize the loss landscape (i.e., the contour line view) of multiple binarization methods before and after implementing ES-BNN. As shown in Figure~\ref{fig:landscape}, ES-BNN noticeably smoothes the rugged and discontinuous surfaces, making them more closely resemble that of the real-valued network. This eventually makes it easier to optimize and escape from sub-optimal convergence, suggesting the advantage of the enhanced representation capacity by our approach.

%% file: sec/6_complexity.tex
\noindent\textbf{Computation Complexity.} 
Finally, we elaborate on the computation complexity and conduct a thorough analysis of the computational impact introduced by ES-BNN. The majority of binarization research commonly employs a unified metric, referred to as OPs. It can be calculated by $\text{OPs} = \text{FLOPs} + \text{BOPs} / 64$, where FLOPs and BOPs denote the number of floating point operations and binary operations. In line with the previous works, we also report the computation complexity of our approach in OPs. However, it is noteworthy that the previous works regard the computation cost of batch normalization as insignificant and omit it from OPs. Yet the computation cost of batch normalization in our approach is increased due to the expanding operation. To better elucidate the computation complexity of our approach, we define a more comprehensive metric OPs+, which takes the computation cost of batch normalization into the overall measurement.    
Table~\ref{t_ops} shows the breakdown of computation complexity in multiple metrics. As we can see, ES-BNN incurs negligible increase in the computation complexity in terms of both OPs (2.2\%) and OPs+ (5.6\%), while achieving significant accuracy improvement (3.4\%).

\begin{table}[t]
\small 
\centering
  \setlength{\tabcolsep}{4pt}
\begin{tabular}{l|ccc}
\toprule
                 &Baseline &ES-BNN &$\Delta$ \\ \midrule\midrule
Conv FLOPs ($10^8$) &1.373     &1.373   &0.000\\
FC FLOPs ($10^8$)   &0.005    &0.041  &0.036 \\
BOPs ($10^9$)       &1.676     &1.676   &0.000     \\
OPs ($10^8$)        &1.640     &1.676   &0.036 (2.2\%) \\  \midrule
BN FLOPs ($10^8$)   &0.025    &0.082  &0.057 \\
OPs+ ($10^8$)      &1.665    &1.758   &0.093 (5.6\%) \\ \midrule 
Top-1       &61.0    &64.4   & 3.4   \\ \bottomrule
\end{tabular}
\caption{Comparison of the computation complexity measured in different metrics (FLOPs, BOPs, OPs and OPs+). We break down the computation into convolution (Conv), fully-connected (FC), and batch normalization (BN) layers. We report the computation complexity and top-1 accuracy (\%) using ResNet-18 on ImageNet.} 
\label{t_ops}
\end{table}

%% file: sec/7_conclusion.tex
\section{Conclusion}

We have presented ES-BNN, which is a simple yet effective approach to improve various BNNs at negligible increase of computation complexity. ES-BNN achieves performance boosts by the proposed expanding-and-shrinking operation to enhance the representation capacity of each binary layer. Extensive experiments on multiple benchmarks reveal that ES-BNN obtains considerable gains over a wide range of binarization algorithms with the architectures of both CNNs and Transformers. Moreover, our approach shows strong generalizability in different applications from image classification, object detection to generative diffusion model.

%% file: sec/X_suppl.tex
\clearpage
\section*{Appendix}
 
Section~\ref{sec:implement} presents more details regarding our implementation. 
Section~\ref{sec:experiment} introduces more details about the multiple benchmark datasets and experimental results. 
Section~\ref{sec:binaryvit} elucidates the application of ES-BNN on BinaryViT. 
Section~\ref{sec:zero_padding} describes more context about the zero padding.

\section{More Implementation Details}
\label{sec:implement}
ES-BNN can be easily implemented in a few lines of code in PyTorch. Following illustrates the python code of integrating the proposed expanding-and-shrinking operation into a standard binary convolution block including binary convolution, batch normalization and residual connection.
\vspace{10pt}
\begin{lstlisting}[language=Python, basicstyle=\ttfamily\fontsize{6.3}{9}\selectfont]
# ichn (ochn): number of input (output) channels
# ctau (ntau): scaling factor ($\tau$) of current (next) layers
# cgrp (ngrp): groups of current (next) layers
# ifeat: \hspace{15pt} input feature maps

class ESBNN_Block(nn.Module):
    def __init__(self, ichn, ochn, ctau, ntau, cgrp, ngrp):
        super(ESBNN_Block, self).__init__()
        self.conv = BConv2d(ichn * ctau * cgrp, int(ochn / ctau), groups = cgrp)
        self.replication = ctau * ntau * ngrp
        self.bn = nn.BatchNorm2d(ochn * ntau * ngrp)
    def forward(self, ifeat):
        ofeat = self.conv(ifeat)
        ofeat = ofeat.repeat(1, self.replication, 1, 1)
        return self.bn(ofeat) + ifeat
\end{lstlisting}

As described in the main paper, all of our experiments are conducted by integrating the proposed expanding-and-shrinking operation into the official open-source codebase of various binarization methods, without changing their original training settings. Table~\ref{table:sources} lists the sources that are used to perform our experiments. 

\begin{table}[b]
\centering
\resizebox{\linewidth}{!}{
  \setlength{\tabcolsep}{4pt}
\begin{tabular}{ll}
\toprule
 Method                &Source \\ \midrule\midrule
Bi-Real~\cite{liu2018birealnet} &\href{https://github.com/liuzechun/Bi-Real-net/}{github.com/liuzechun/Bi-Real-net}    \\
IR-Net~\cite{qin2020irnet} &\href{https://github.com/htqin/IR-Net/}{github.com/htqin/IR-Net}    \\
ReCU~\cite{xu2021recu} &\href{https://github.com/z-hXu/ReCU/}{github.com/z-hXu/ReCU}    \\
RBNN~\cite{lin2020rbnn} &\href{https://github.com/lmbxmu/RBNN/}{github.com/lmbxmu/RBNN}    \\
AdaBin~\cite{tu2022adabin} &\href{https://github.com/huawei-noah/Efficient-Computing/tree/master/BinaryNetworks/AdaBin/}{github.com/huawei-noah/Efficient-Computing}    \\
SiMaN~\cite{lin2022siman} &\href{https://github.com/lmbxmu/SiMaN/}{github.com/lmbxmu/SiMaN}    \\
ReActNet~\cite{liu2020reactnet} &\href{https://github.com/liuzechun/ReActNet/}{github.com/liuzechun/ReActNet}    \\
BNext~\cite{guo2022bnext} &\href{https://github.com/hpi-xnor/BNext}{github.com/hpi-xnor/BNext}    \\
BinaryViT~\cite{le2023binaryvit} &\href{https://github.com/Phuoc-Hoan-Le/BinaryViT}{github.com/Phuoc-Hoan-Le/BinaryViT}    \\
BiDet~\cite{wang2020bidet} &\href{https://github.com/ZiweiWangTHU/BiDet}{github.com/ZiweiWangTHU/BiDet}    \\
BI-DiffSR~\cite{chen2024binarized} &\href{https://github.com/zhengchen1999/BI-DiffSR}{github.com/zhengchen1999/BI-DiffSR}    \\  \bottomrule
\end{tabular}
}
\caption{Sources of different methods used in our experiments.} 
\label{table:sources}
\vspace{-2mm}
\end{table}

\section{More Experimental Details}
\label{sec:experiment}
\noindent\textbf{Datasets.} 
We conduct extensive experiments on a broad range of datasets, including CIFAR-10~\cite{krizhevsky2009cifar10}, ImageNet~\cite{russakovsky2015imagenet}, PASCAL VOC~\cite{mark2010voc}, DIV2K~\cite{timofte2017ntire}, Flickr2K~\cite{bevilacqua2012low}, Set5~\cite{martin2001database}, B100~\cite{huang2015single}, Urban100~\cite{huang2015single} and Manga109~\cite{matsui2017sketch}. 
CIFAR-10 is widely used in the binarization research, and consists of 60,000 images in 10 categories. It comprises 50,000 images for training and 10,000 images for testing.
ImageNet is a large-scale dataset of 1,000 categories including 1.2M training images and 50K validation images. PASCAL VOC covers 20 categories and owns two versions, i.e., VOC-2007 and VOC-2012. To be consistent with the previous methods~\cite{wang2020bidet}, we combine the training sets from both versions, totaling 16,000 images for training, and the test set from VOC 2007 is used. 
DIV2K and Flickr2K consist of 1,000 and 2,650 images, respectively, and are commonly utilized as training datasets for image super-resolution. For evaluation, Set5, B100, Urban100 and Manga109 serve as widely adopted benchmarks. Set5 is a compact dataset of 5 images, often used for quick assessments. B100, derived from the Berkeley segmentation dataset~\cite{huang2015single}, includes 100 images featuring diverse content. Urban100 contains 100 images of urban scenes, capturing the intricate details of urban environments. Manga109 is distinct, comprising 109 manga images, offering a unique challenge due to the presence of line art and text characteristic of manga.

\begin{table*}[t]
  \centering
  \resizebox{\linewidth}{!}{
\begin{tabular}{c|cccccccccccccccc|cccccccccccccccc}
\hline
      & \multicolumn{16}{c|}{$\tau$}                                                & \multicolumn{16}{c}{$g$}                                                \\ \midrule \midrule
Layer & 1 & 2 & 3 & 4 & 5 & 6 & 7 & 8 & 9 & 10 & 11 & 12 & 13 & 14 & 15 & 16 & 1 & 2 & 3 & 4 & 5 & 6 & 7 & 8 & 9 & 10 & 11 & 12 & 13 & 14 & 15 & 16 \\
\midrule
Param. & 2 & 2 & 2 & 2 & 4 & 4 & 2 & 2 & 4 & 4  & 4  & 4  & 4  & 4  & 4  & 4  & 1 & 1 & 2 & 2 & 1 & 1 & 2 & 2 & 1 & 1  & 1  & 1  & 1  & 1  & 2  & 2  \\
\bottomrule
\end{tabular}
}
  \caption{Illustration of the setting of hyper-parameters $\tau$ and $g$.}
  \label{table:hyperparameter_setting}
\end{table*}

\begin{figure}[t]
   \centering
   \includegraphics[width=\linewidth]{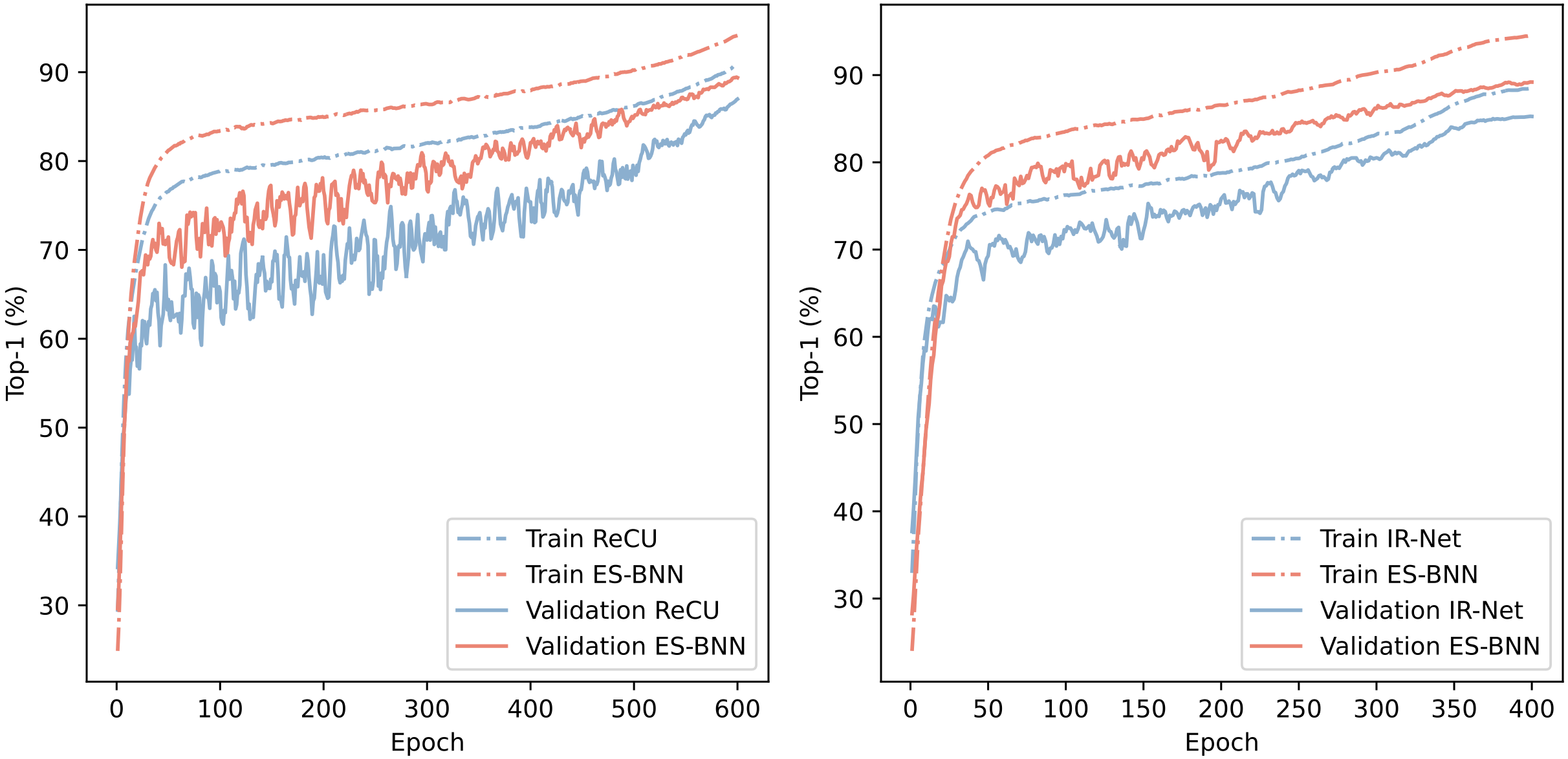}
   \caption{Comparison of the learning behaviors between the two baseline methods (ReCU and IR-Net) and the proposed approach using ResNet-20 on CIFAR-10.}
   \label{fig:training_curves}
\end{figure}

\noindent\textbf{Learning Behaviors.} 
To gain the insight into how ES-BNN impacts on the learning behavior of a binary network, we provide the training process of the baseline methods and our approach. As demonstrated in Figure~\ref{fig:training_curves}, it is observed that ES-BNN yields steady improvement in both training and validation accuracy throughout the learning process. Moreover, this improving trend is consistent across different binarization methods used as the baselines.

\begin{table}[t]
  \centering
  \resizebox{\linewidth}{!}{
  \begin{tabular}{l|c|ccc}
    \toprule
  \multirow{2}{*}{Method}     & \multirow{2}{*}{Baseline ($n$)}        &\multicolumn{3}{c}{ES-BNN}    \\ \cline{3-5} 
       &          & $n\tau$ & $n$ & $n/\tau$    \\
    \midrule \midrule
   Bi-Real~\cite{liu2018birealnet}  & 81.4          & 83.4 (+2.0)  & 83.5 (+2.1) &  83.4 (+2.0)    \\ 
   IR-Net~\cite{qin2020irnet}       & 86.5              & 88.7 (+2.2) & 88.5 (+2.0) & 88.4 (+1.9)\\
   ReCU~\cite{xu2021recu}         & 87.5                & 89.0 (+1.5)  & 89.1 (+1.6) & 88.8 (+1.3) \\
   RBNN~\cite{lin2020rbnn}         & 87.8                & 89.4 (+1.6) &89.3 (+1.5)  &89.1 (+1.3)\\
   SiMaN~\cite{lin2022siman}         & 87.4                & 89.2 (+1.8)     &89.1 (+1.7)  &88.9 (+1.5)\\   
   AdaBin~\cite{tu2022adabin}        & 88.2                 & 89.5 (+1.3)      &89.4 (+1.2)    &89.3 (+1.1)\\          
    \bottomrule
  \end{tabular}
  }
  \caption{Comparison of the top-1 accuracy (\%) of the baseline methods and our approach using different channel numbers in the last fully connected layer.}
  \label{table:fc_type}
\end{table}

\begin{figure}
    \centering
    \includegraphics[width = \linewidth]{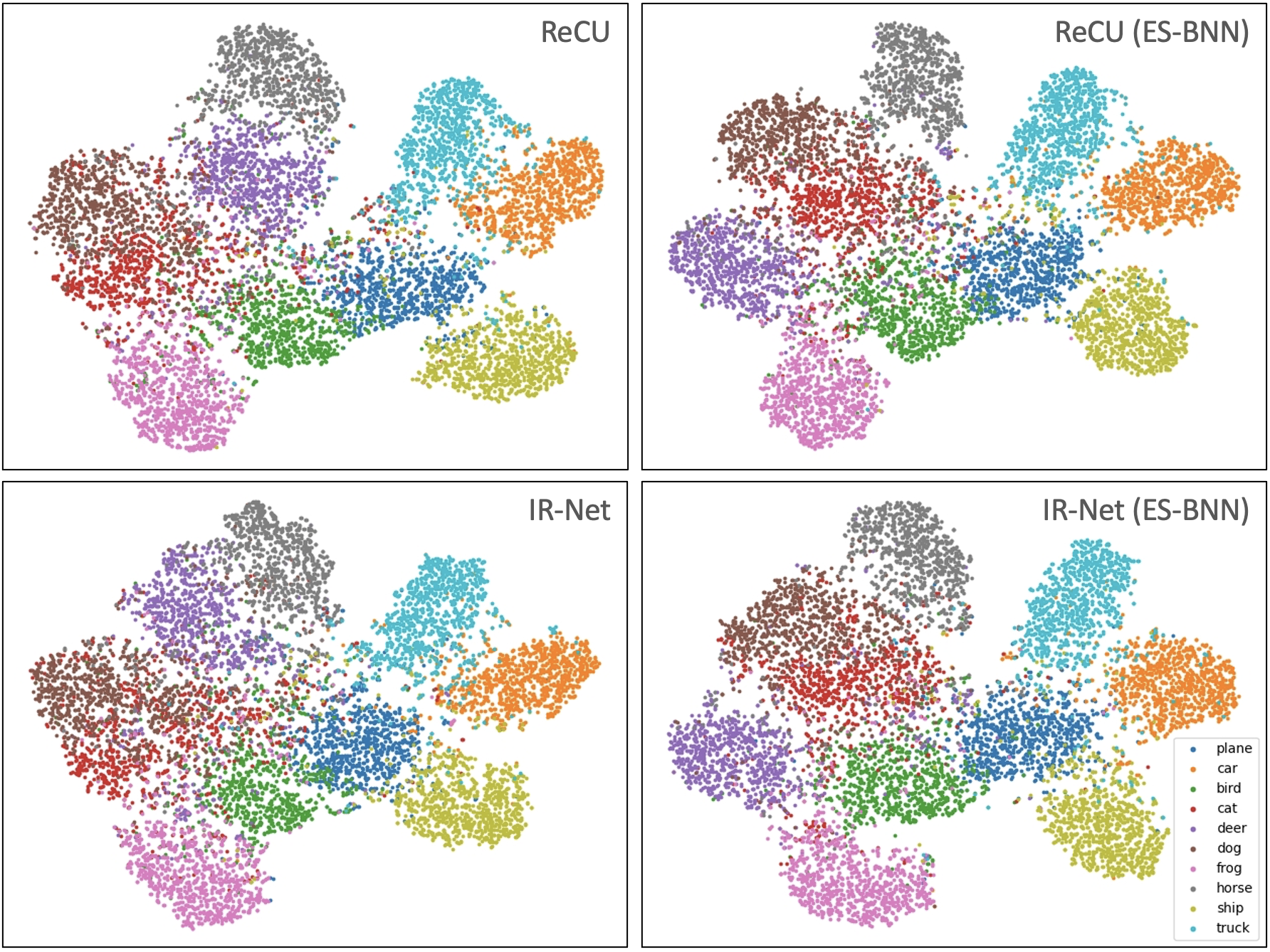}
    \caption{t-SNE visualization of the features extracted before the last fully-connected layer of baseline methods and our approach using ResNet-20 on CIFAR-10.}
    \label{fig:tsne}
\end{figure}

\noindent\textbf{Feature Visualization.} 
We further analyze the feature distribution for a deeper comprehension of the enhancement achieved by ES-BNN. Figure~\ref{fig:tsne} visualizes the features extracted before the last fully-connected layer of the baselines and our approach using ResNet-20 on CIFAR-10. It can be found that the inter-class features of our approach are overall more scattered. For instance, distinguishing truck (cyan) and car (orange) in ReCU, as well as between deer (purple) and horse (gray) in IR-Net, which is relatively challenging in the baselines, is improved in ES-BNN.   

\noindent\textbf{Fully Connected Layer.}
In the main paper, we keep using the replication for the last fully connected layer due to its negligible computation overhead. In practice, the accuracy can be still improved by keeping the number of channels in this layer unchanged or reduced. Table~\ref{table:fc_type} shows the three cases in the last fully connected layer: 1) the same replication as used in convolution layers, i.e., $n\tau$ channels, 2) the number of channels is simply replicated to match the original one, i.e., $n$ channels, 3) no replication, i.e., $n / \tau$ channels. We experiment on multiple baseline methods under the setting of $\tau = 2$ and $g = 1$ with ResNet-20 on CIFAR-10. As can be seen in the table, ES-BNN is overall insensitive to the number of channels in the last fully connected layer. This implies that our approach can still bring significant improvement even with lower OPs than the baseline methods.

\noindent\textbf{Hyper-Parameters}
As introduced in Section 3.4 of the main paper, we follow the principle of general network design to use larger $\tau g$ for the deeper layers so as to enhance the representation capacity more in the deeper layers for feature abstraction. Table~\ref{table:hyperparameter_setting} shows the specific $\tau$ and $g$ in each layer of ResNet-18.

\section{More Details on BinaryViT}
\label{sec:binaryvit}
\begin{figure}[h]
\centering
\includegraphics[width=0.8\linewidth]{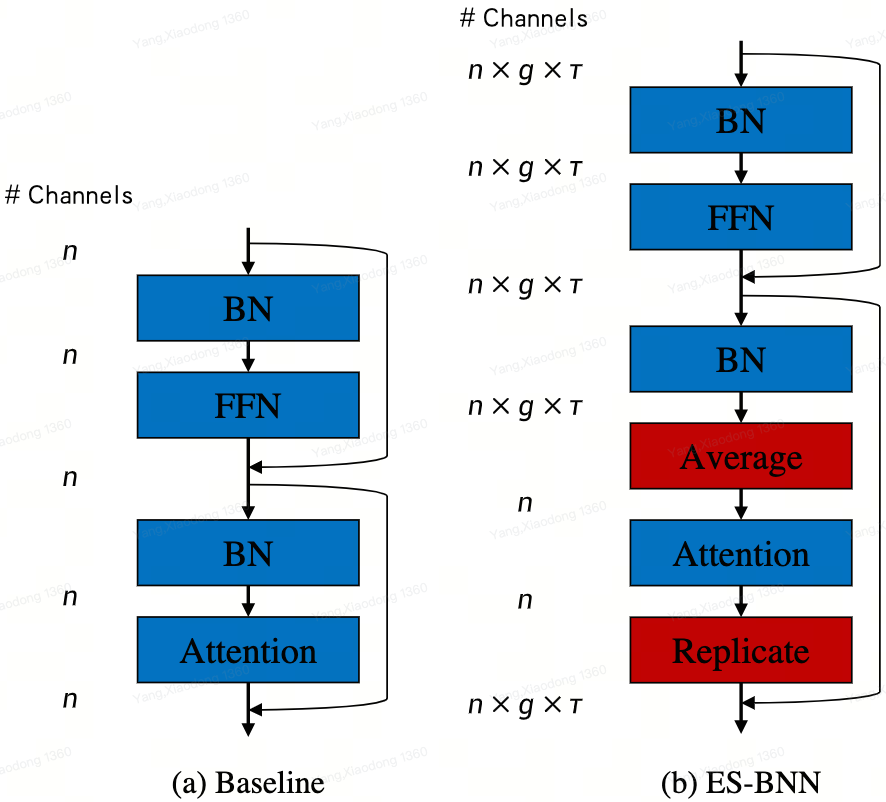}
\caption{Illustration of how ES-BNN is applied to BinaryViT.}
\label{fig:binaryvit}
\end{figure}

To provide more details of how ES-BNN is applied to BinaryViT, we present a schematic diagram in Figure~\ref{fig:binaryvit}. For the feed-forward network (FFN), we directly apply our approach to increase its representation capacity. For the attention layers (both self-attention and cross-attention), we take average of the features to revert to the original feature dimension, thereby maintaining the initial complexity of the attention layers.

\section{Zero Padding}
\label{sec:zero_padding}

\begin{figure}[t]
\centering
\includegraphics[width=\linewidth]{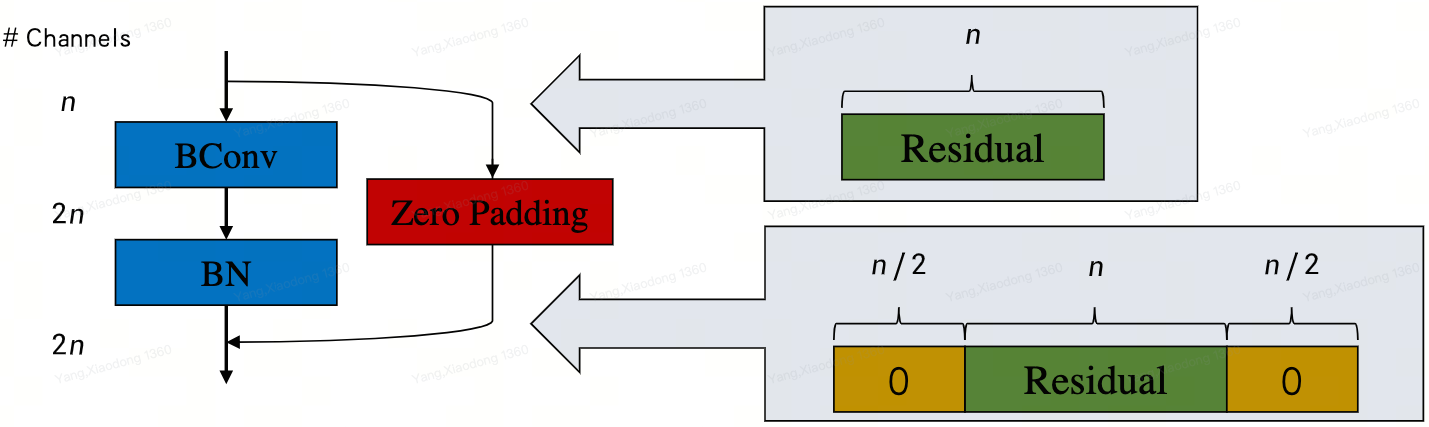}
\caption{Illustration of zero padding of residuals in ResNet-20.}
\label{fig:zero_padding}
\end{figure}

As shown in Figure 7 of the main paper, the abrupt shifts at the 9th and 15th layers are due to the zero-padding in feature channels of residual connections at the corresponding layers in ResNet-20. Here we provide more context. Traditionally, zero-padding is used in the spatial dimensions to preserve the input feature map size. However, in ResNet-20, zero-padding is used in the feature dimension to compensate for the feature dimension mismatch of residual connections in the 9th and 15th layers, as illustrated in Figure~\ref{fig:zero_padding}. As a result of the padded zeros, the correlation coefficients of the two layers shift dramatically.

%% file: main.bbl
\begin{thebibliography}{55}
\providecommand{\natexlab}[1]{#1}
\providecommand{\url}[1]{\texttt{#1}}
\expandafter\ifx\csname urlstyle\endcsname\relax
  \providecommand{\doi}[1]{doi: #1}\else
  \providecommand{\doi}{doi: \begingroup \urlstyle{rm}\Url}\fi

\bibitem[Ayad et~al.(2024)Ayad, Larue1, and Nguyen]{qn-mixer}
Ishak Ayad, Nicolas Larue1, and Mai Nguyen.
\newblock Qn-mixer: A quasi-newton mlp-mixer model for sparse-view ct reconstruction.
\newblock In \emph{CVPR}, 2024.

\bibitem[Bethge et~al.(2021)Bethge, Bartz, Yang, Chen, and Meinel]{bethge2020meliusnet}
Joseph Bethge, Christian Bartz, Haojin Yang, Ying Chen, and Christoph Meinel.
\newblock Meliusnet: An improved network architecture for binary neural networks.
\newblock In \emph{WACV}, 2021.

\bibitem[Bevilacqua et~al.(2012)Bevilacqua, Roumy, Guillemot, and Alberi-Morel]{bevilacqua2012low}
Marco Bevilacqua, Aline Roumy, Christine Guillemot, and Marie~Line Alberi-Morel.
\newblock Low-complexity single-image super-resolution based on nonnegative neighbor embedding.
\newblock 2012.

\bibitem[Bulat and Tzimiropoulos(2019)]{bulat2019xnor++}
Adrian Bulat and Georgios Tzimiropoulos.
\newblock Xnor-net++: Improved binary neural networks.
\newblock \emph{arXiv preprint arXiv:1909.13863}, 2019.

\bibitem[Chen et~al.(2024)Chen, Qin, Guo, Su, Yuan, Kong, and Zhang]{chen2024binarized}
Zheng Chen, Haotong Qin, Yong Guo, Xiongfei Su, Xin Yuan, Linghe Kong, and Yulun Zhang.
\newblock Binarized diffusion model for image super-resolution.
\newblock In \emph{NeurIPS}, 2024.

\bibitem[Das et~al.(2024)Das, Wewer, Yunus, Ilg, and Lenssen]{recon}
Devikalyan Das, Christopher Wewer, Raza Yunus, Eddy Ilg, and Jan~Eric Lenssen.
\newblock Neural parametric gaussians for monocular non-rigid object reconstruction.
\newblock In \emph{CVPR}, 2024.

\bibitem[Esser et~al.(2020)Esser, McKinstry, Bablani, Appuswamy, and Modha]{esser2019lsq}
Steven~K Esser, Jeffrey~L McKinstry, Deepika Bablani, Rathinakumar Appuswamy, and Dharmendra~S Modha.
\newblock Learned step size quantization.
\newblock In \emph{ICLR}, 2020.

\bibitem[Everingham et~al.(2010)Everingham, {van Gool}, Williams, Winn, and Zisserman]{mark2010voc}
Mark Everingham, Luc {van Gool}, {Christopher K. I.} Williams, John Winn, and Andrew Zisserman.
\newblock The pascal visual object classes (voc) challenge.
\newblock \emph{International Journal of Computer Vision}, 88\penalty0 (2):\penalty0 303--338, 2010.

\bibitem[Gong et~al.(2019)Gong, Liu, Jiang, Li, Hu, Lin, Yu, and Yan]{gong2019dsq}
Ruihao Gong, Xianglong Liu, Shenghu Jiang, Tianxiang Li, Peng Hu, Jiazhen Lin, Fengwei Yu, and Junjie Yan.
\newblock Differentiable soft quantization: Bridging full-precision and low-bit neural networks.
\newblock In \emph{ICCV}, 2019.

\bibitem[Guo et~al.(2022)Guo, Bethge, Meinel, and Yang]{guo2022bnext}
Nianhui Guo, Joseph Bethge, Christoph Meinel, and Haojin Yang.
\newblock Join the high accuracy club on imagenet with a binary neural network ticket.
\newblock \emph{arXiv preprint arXiv:2211.12933}, 2022.

\bibitem[Hinton et~al.(2015)Hinton, Vinyals, and Dean]{hinton2015distilling}
Geoffrey Hinton, Oriol Vinyals, and Jeff Dean.
\newblock Distilling the knowledge in a neural network.
\newblock \emph{arXiv preprint arXiv:1503.02531}, 2015.

\bibitem[Ho et~al.(2020)Ho, Jain, and Abbeel]{dm1}
Jonathan Ho, Ajay Jain, and Pieter Abbeel.
\newblock Denoising diffusion probabilistic models.
\newblock In \emph{NeurIPS}, 2020.

\bibitem[Howard et~al.(2017)Howard, Zhu, Chen, Kalenichenko, Wang, Weyand, Andreetto, and Adam]{howard2017mobilenets}
Andrew~G Howard, Menglong Zhu, Bo Chen, Dmitry Kalenichenko, Weijun Wang, Tobias Weyand, Marco Andreetto, and Hartwig Adam.
\newblock Mobilenets: Efficient convolutional neural networks for mobile vision applications.
\newblock \emph{arXiv preprint arXiv:1704.04861}, 2017.

\bibitem[Huang et~al.(2015)Huang, Singh, and Ahuja]{huang2015single}
Jia-Bin Huang, Abhishek Singh, and Narendra Ahuja.
\newblock Single image super-resolution from transformed self-exemplars.
\newblock In \emph{Proceedings of the IEEE conference on computer vision and pattern recognition}, pages 5197--5206, 2015.

\bibitem[Hubara et~al.(2016)Hubara, Courbariaux, Soudry, El-Yaniv, and Bengio]{courbariaux2016bnns}
Itay Hubara, Matthieu Courbariaux, Daniel Soudry, Ran El-Yaniv, and Yoshua Bengio.
\newblock Binarized neural networks.
\newblock \emph{Advances in neural information processing systems}, 29, 2016.

\bibitem[Krizhevsky et~al.(2009)Krizhevsky, Hinton, et~al.]{krizhevsky2009cifar10}
Alex Krizhevsky, Geoffrey Hinton, et~al.
\newblock Learning multiple layers of features from tiny images.
\newblock 2009.

\bibitem[Le and Li(2023)]{le2023binaryvit}
Phuoc-Hoan~Charles Le and Xinlin Li.
\newblock Binaryvit: Pushing binary vision transformers towards convolutional models.
\newblock In \emph{Proceedings of the IEEE/CVF Conference on Computer Vision and Pattern Recognition}, pages 4664--4673, 2023.

\bibitem[Lee et~al.(2023)Lee, Kim, Park, and Kim]{lee2023insta}
Changhun Lee, Hyungjun Kim, Eunhyeok Park, and Jae-Joon Kim.
\newblock Insta-bnn: Binary neural network with instance-aware threshold.
\newblock In \emph{Proceedings of the IEEE/CVF International Conference on Computer Vision}, pages 17325--17334, 2023.

\bibitem[Li et~al.(2023)Li, Luo, and Yang]{pillarnext}
Jinyu Li, Chenxu Luo, and Xiaodong Yang.
\newblock Pillarnext: Rethinking network designs for 3d object detection in lidar point clouds.
\newblock In \emph{CVPR}, 2023.

\bibitem[Li and Yang(2023)]{tip}
Weixin Li and Xiaodong Yang.
\newblock Transcendental idealism of planner: Evaluating perception from planning perspective for autonomous driving.
\newblock In \emph{ICML}, 2023.

\bibitem[Li et~al.(2022)Li, Adamczewski, Li, Gu, Timofte, and Van~Gool]{li2022randompruning}
Yawei Li, Kamil Adamczewski, Wen Li, Shuhang Gu, Radu Timofte, and Luc Van~Gool.
\newblock Revisiting random channel pruning for neural network compression.
\newblock In \emph{Proceedings of the IEEE/CVF Conference on Computer Vision and Pattern Recognition}, pages 191--201, 2022.

\bibitem[Lin et~al.(2020)Lin, Ji, Xu, Zhang, Wang, Wu, Huang, and Lin]{lin2020rbnn}
Mingbao Lin, Rongrong Ji, Zihan Xu, Baochang Zhang, Yan Wang, Yongjian Wu, Feiyue Huang, and Chia-Wen Lin.
\newblock Rotated binary neural network.
\newblock \emph{Advances in neural information processing systems}, 33:\penalty0 7474--7485, 2020.

\bibitem[Lin et~al.(2022)Lin, Ji, Xu, Zhang, Chao, Lin, and Shao]{lin2022siman}
Mingbao Lin, Rongrong Ji, Zihan Xu, Baochang Zhang, Fei Chao, Chia-Wen Lin, and Ling Shao.
\newblock Siman: Sign-to-magnitude network binarization.
\newblock \emph{IEEE Transactions on Pattern Analysis and Machine Intelligence}, 45\penalty0 (5):\penalty0 6277--6288, 2022.

\bibitem[Liu et~al.(2018{\natexlab{a}})Liu, Simonyan, and Yang]{liu2018darts}
Hanxiao Liu, Karen Simonyan, and Yiming Yang.
\newblock Darts: Differentiable architecture search.
\newblock In \emph{International Conference on Learning Representations}, 2018{\natexlab{a}}.

\bibitem[Liu and Kwon(2025)]{ar}
Yongfan Liu and Hyoukjun Kwon.
\newblock Efficient stereo depth estimation model for wearable augmented reality devices.
\newblock In \emph{CVPR}, 2025.

\bibitem[Liu et~al.(2017)Liu, Li, Shen, Huang, Yan, and Zhang]{liu2017slimming}
Zhuang Liu, Jianguo Li, Zhiqiang Shen, Gao Huang, Shoumeng Yan, and Changshui Zhang.
\newblock Learning efficient convolutional networks through network slimming.
\newblock In \emph{Proceedings of the IEEE international conference on computer vision}, pages 2736--2744, 2017.

\bibitem[Liu et~al.(2018{\natexlab{b}})Liu, Wu, Luo, Yang, Liu, and Cheng]{liu2018birealnet}
Zechun Liu, Baoyuan Wu, Wenhan Luo, Xin Yang, Wei Liu, and Kwang-Ting Cheng.
\newblock Bi-real net: Enhancing the performance of 1-bit cnns with improved representational capability and advanced training algorithm.
\newblock In \emph{Proceedings of the European conference on computer vision (ECCV)}, pages 722--737, 2018{\natexlab{b}}.

\bibitem[Liu et~al.(2019)Liu, Mu, Zhang, Guo, Yang, Cheng, and Sun]{liu2019metapruning}
Zechun Liu, Haoyuan Mu, Xiangyu Zhang, Zichao Guo, Xin Yang, Kwang-Ting Cheng, and Jian Sun.
\newblock Metapruning: Meta learning for automatic neural network channel pruning.
\newblock In \emph{Proceedings of the IEEE/CVF international conference on computer vision}, pages 3296--3305, 2019.

\bibitem[Liu et~al.(2020)Liu, Shen, Savvides, and Cheng]{liu2020reactnet}
Zechun Liu, Zhiqiang Shen, Marios Savvides, and Kwang-Ting Cheng.
\newblock Reactnet: Towards precise binary neural network with generalized activation functions.
\newblock In \emph{European Conference on Computer Vision}, pages 143--159. Springer, 2020.

\bibitem[Liu et~al.(2021)Liu, Shen, Li, Helwegen, Huang, and Cheng]{liu2021adambnn}
Zechun Liu, Zhiqiang Shen, Shichao Li, Koen Helwegen, Dong Huang, and Kwang-Ting Cheng.
\newblock How do adam and training strategies help bnns optimization.
\newblock In \emph{International conference on machine learning}, pages 6936--6946. PMLR, 2021.

\bibitem[Martin et~al.(2001)Martin, Fowlkes, Tal, and Malik]{martin2001database}
David Martin, Charless Fowlkes, Doron Tal, and Jitendra Malik.
\newblock A database of human segmented natural images and its application to evaluating segmentation algorithms and measuring ecological statistics.
\newblock In \emph{Proceedings eighth IEEE international conference on computer vision. ICCV 2001}, pages 416--423. IEEE, 2001.

\bibitem[Martinez et~al.(2020)Martinez, Yang, Bulat, and Tzimiropoulos]{martinez2020real-to-binary}
Brais Martinez, Jing Yang, Adrian Bulat, and Georgios Tzimiropoulos.
\newblock Training binary neural networks with real-to-binary convolutions.
\newblock In \emph{International Conference on Learning Representations}, 2020.

\bibitem[Matsui et~al.(2017)Matsui, Ito, Aramaki, Fujimoto, Ogawa, Yamasaki, and Aizawa]{matsui2017sketch}
Yusuke Matsui, Kota Ito, Yuji Aramaki, Azuma Fujimoto, Toru Ogawa, Toshihiko Yamasaki, and Kiyoharu Aizawa.
\newblock Sketch-based manga retrieval using manga109 dataset.
\newblock \emph{Multimedia tools and applications}, 76:\penalty0 21811--21838, 2017.

\bibitem[Nguyen et~al.(2024)Nguyen, Nguyen, and Luu]{video}
Trong-Thuan Nguyen, Pha Nguyen, and Khoa Luu.
\newblock Hig: Hierarchical interlacement graph approach to scene graph generation in video understanding.
\newblock In \emph{CVPR}, 2024.

\bibitem[Qin et~al.(2020)Qin, Gong, Liu, Shen, Wei, Yu, and Song]{qin2020irnet}
Haotong Qin, Ruihao Gong, Xianglong Liu, Mingzhu Shen, Ziran Wei, Fengwei Yu, and Jingkuan Song.
\newblock Forward and backward information retention for accurate binary neural networks.
\newblock In \emph{Proceedings of the IEEE/CVF Conference on Computer Vision and Pattern Recognition}, pages 2250--2259, 2020.

\bibitem[Rastegari et~al.(2016)Rastegari, Ordonez, Redmon, and Farhadi]{rastegari2016xnor}
Mohammad Rastegari, Vicente Ordonez, Joseph Redmon, and Ali Farhadi.
\newblock Xnor-net: Imagenet classification using binary convolutional neural networks.
\newblock In \emph{European conference on computer vision}, pages 525--542. Springer, 2016.

\bibitem[Rombach et~al.(2022)Rombach, Blattmann, Lorenz, Esser, and Ommer]{dm2}
Robin Rombach, Andreas Blattmann, Dominik Lorenz, Patrick Esser, and Bjorn Ommer.
\newblock High-resolution image synthesis with latent diffusion models.
\newblock In \emph{CVPR}, 2022.

\bibitem[Russakovsky et~al.(2015)Russakovsky, Deng, Su, Krause, Satheesh, Ma, Huang, Karpathy, Khosla, Bernstein, et~al.]{russakovsky2015imagenet}
Olga Russakovsky, Jia Deng, Hao Su, Jonathan Krause, Sanjeev Satheesh, Sean Ma, Zhiheng Huang, Andrej Karpathy, Aditya Khosla, Michael Bernstein, et~al.
\newblock Imagenet large scale visual recognition challenge.
\newblock \emph{International journal of computer vision}, 115\penalty0 (3):\penalty0 211--252, 2015.

\bibitem[Smith et~al.(2024)Smith, Cao, and Levine]{x-embodiment}
Laura Smith, Yunhao Cao, and Sergey Levine.
\newblock Grow your limits: Continuous improvement with real-world rl for robotic locomotion.
\newblock In \emph{ICRA}, 2024.

\bibitem[Soudry et~al.(2014)Soudry, Hubara, and Meir]{ebp}
Daniel Soudry, Itay Hubara, and Ron Meir.
\newblock Expectation backpropagation: Parameter-free training of multilayer neural networks with continuous or discrete weights.
\newblock In \emph{NeurIPS}, 2014.

\bibitem[Timofte et~al.(2017)Timofte, Agustsson, Van~Gool, Yang, and Zhang]{timofte2017ntire}
Radu Timofte, Eirikur Agustsson, Luc Van~Gool, Ming-Hsuan Yang, and Lei Zhang.
\newblock Ntire 2017 challenge on single image super-resolution: Methods and results.
\newblock In \emph{Proceedings of the IEEE conference on computer vision and pattern recognition workshops}, pages 114--125, 2017.

\bibitem[Tu et~al.(2022)Tu, Chen, Ren, and Wang]{tu2022adabin}
Zhijun Tu, Xinghao Chen, Pengju Ren, and Yunhe Wang.
\newblock Adabin: Improving binary neural networks with adaptive binary sets.
\newblock In \emph{European conference on computer vision}, pages 379--395. Springer, 2022.

\bibitem[Wang et~al.(2020)Wang, Wu, Lu, and Zhou]{wang2020bidet}
Ziwei Wang, Ziyi Wu, Jiwen Lu, and Jie Zhou.
\newblock Bidet: An efficient binarized object detector.
\newblock In \emph{Proceedings of the IEEE/CVF conference on computer vision and pattern recognition}, pages 2049--2058, 2020.

\bibitem[Wang et~al.(2021)Wang, Lu, Wu, and Zhou]{wang2021autobidet}
Ziwei Wang, Jiwen Lu, Ziyi Wu, and Jie Zhou.
\newblock Learning efficient binarized object detectors with information compression.
\newblock \emph{IEEE Transactions on Pattern Analysis and Machine Intelligence}, 44\penalty0 (6):\penalty0 3082--3095, 2021.

\bibitem[Wang et~al.(2023)Wang, Li, Luo, Xie, and Yang]{wang2023distillbev}
Zeyu Wang, Dingwen Li, Chenxu Luo, Cihang Xie, and Xiaodong Yang.
\newblock Distillbev: Boosting multi-camera 3d object detection with cross-modal knowledge distillation.
\newblock In \emph{Proceedings of the IEEE/CVF International Conference on Computer Vision}, pages 8637--8646, 2023.

\bibitem[Wu et~al.(2023)Wu, Zheng, Liu, and Zheng]{wu2023reste}
Xiao-Ming Wu, Dian Zheng, Zuhao Liu, and Wei-Shi Zheng.
\newblock Estimator meets equilibrium perspective: A rectified straight through estimator for binary neural networks training.
\newblock In \emph{Proceedings of the IEEE/CVF International Conference on Computer Vision}, pages 17055--17064, 2023.

\bibitem[Xia et~al.(2023)Xia, Zhang, Wang, Tian, Yang, Timofte, and Van~Gool]{xia2022bbcu}
Bin Xia, Yulun Zhang, Yitong Wang, Yapeng Tian, Wenming Yang, Radu Timofte, and Luc Van~Gool.
\newblock Basic binary convolution unit for binarized image restoration network.
\newblock \emph{ICLR}, 2023.

\bibitem[Xu et~al.(2021)Xu, Lin, Liu, Chen, Shao, Gao, Tian, and Ji]{xu2021recu}
Zihan Xu, Mingbao Lin, Jianzhuang Liu, Jie Chen, Ling Shao, Yue Gao, Yonghong Tian, and Rongrong Ji.
\newblock Recu: Reviving the dead weights in binary neural networks.
\newblock In \emph{Proceedings of the IEEE/CVF international conference on computer vision}, pages 5198--5208, 2021.

\bibitem[Yang et~al.(2018)Yang, Molchanov, and Kautz]{prernn}
Xiaodong Yang, Pavlo Molchanov, and Jan Kautz.
\newblock Making convolutional networks recurrent for visual sequence learning.
\newblock In \emph{CVPR}, 2018.

\bibitem[Yang et~al.(2023)Yang, Ma, Ji, and Ren]{gedepth}
Xiaodong Yang, Zhuang Ma, Zhiyu Ji, and Zhe Ren.
\newblock Gedepth: Ground embedding for monocular depth estimation.
\newblock In \emph{ICCV}, 2023.

\bibitem[Yang et~al.(2022)Yang, Li, Jiang, Gong, Yuan, Zhao, and Yuan]{yang2022focaldistill}
Zhendong Yang, Zhe Li, Xiaohu Jiang, Yuan Gong, Zehuan Yuan, Danpei Zhao, and Chun Yuan.
\newblock Focal and global knowledge distillation for detectors.
\newblock In \emph{Proceedings of the IEEE/CVF Conference on Computer Vision and Pattern Recognition}, pages 4643--4652, 2022.

\bibitem[Zhang et~al.(2018)Zhang, Zhou, Lin, and Sun]{zhang2018shufflenet}
Xiangyu Zhang, Xinyu Zhou, Mengxiao Lin, and Jian Sun.
\newblock Shufflenet: An extremely efficient convolutional neural network for mobile devices.
\newblock In \emph{Proceedings of the IEEE conference on computer vision and pattern recognition}, pages 6848--6856, 2018.

\bibitem[Zhou et~al.(2016)Zhou, Wu, Ni, Zhou, Wen, and Zou]{zhou2016dorefa}
Shuchang Zhou, Yuxin Wu, Zekun Ni, Xinyu Zhou, He Wen, and Yuheng Zou.
\newblock Dorefa-net: Training low bitwidth convolutional neural networks with low bitwidth gradients.
\newblock \emph{arXiv preprint arXiv:1606.06160}, 2016.

\bibitem[Zhu et~al.(2019)Zhu, Dong, and Su]{zhu2019benn}
Shilin Zhu, Xin Dong, and Hao Su.
\newblock Binary ensemble neural network: More bits per network or more networks per bit?
\newblock In \emph{Proceedings of the IEEE/CVF Conference on Computer Vision and Pattern Recognition}, pages 4923--4932, 2019.

\bibitem[Zhuang et~al.(2019)Zhuang, Shen, Tan, Liu, and Reid]{zhuang2019Group-Net}
Bohan Zhuang, Chunhua Shen, Mingkui Tan, Lingqiao Liu, and Ian Reid.
\newblock Structured binary neural networks for accurate image classification and semantic segmentation.
\newblock In \emph{Proceedings of the IEEE/CVF Conference on Computer Vision and Pattern Recognition}, pages 413--422, 2019.

\end{thebibliography}
